\documentclass[journal]{IEEEtran}
\usepackage{cite}
\usepackage{amsmath,amssymb,amsfonts}
\usepackage{algorithmic}
\usepackage{comment}
\usepackage{graphicx}
\usepackage{textcomp}
\usepackage{subfigure}
\usepackage[colorlinks]{hyperref}
\usepackage{multirow}
\usepackage{bbm}

\usepackage[switch]{lineno}
\usepackage{lipsum}  

\def\BibTeX{{\rm B\kern-.05em{\sc i\kern-.025em b}\kern-.08em
    T\kern-.1667em\lower.7ex\hbox{E}\kern-.125emX}}
\begin{document}
\title{VLM-CPL: Consensus pseudo-labels from Vision-Language Models for Annotation-Free Pathological Image Classification}
\author{Lanfeng~Zhong, Zongyao~Huang, Yang~Liu, Wenjun~Liao, Shichuan~Zhang, Guotai~Wang, and Shaoting~Zhang
\thanks{This work was supported by the National Natural Science Foundation of China (62271115). Corresponding author: G. Wang (guotai.wang@uestc.edu.cn)}
\thanks{This work has been submitted to the IEEE for possible publication. Copyright may be transferred without notice, after which this version may no longer be accessible.}
\thanks{Lanfeng~Zhong, Guotai~Wang and Shaoting~Zhang are with the School of Mechanical and Electrical Engineering, University of Electronic Science and Technology of China, Chengdu, 611731, China and also with Shanghai Artificial Intelligence Laboratory, Shanghai 200030, China.}
\thanks{Zongyao~Huang and Yang~Liu are with Department of Pathology, Sichuan Clinical Research Center for
Cancer, Sichuan Cancer Hospital \& Institute, Affiliated Cancer Hospital of University of Electronic
Science and Technology of China, Chengdu, 610042, China.}
\thanks{Wenjun~Liao and Shichuan~Zhang are with Department of Radiation Oncology, Sichuan Cancer Hospital and Institute, University of Electronic Science and Technology of China, Chengdu 610042, China.}
}

\maketitle

\begin{abstract}
Classification of pathological
images is the basis for automatic cancer diagnosis.
Despite that deep learning methods have achieved remarkable 
performance, they heavily 
rely on labeled data, demanding extensive human annotation 
efforts.
In this study, we present a novel human annotation-free method
by leveraging pre-trained Vision-Language Models (VLMs).
Without human annotation, pseudo-labels of the training set are 
obtained by utilizing the zero-shot inference 
capabilities of VLM, which may contain a lot of noise due to the 
domain gap between the pre-training and target datasets. 
To address this issue, 
we introduce VLM-CPL, a novel approach that contains two noisy label filtering techniques with a 
semi-supervised learning strategy.
Specifically, we first obtain prompt-based pseudo-labels with uncertainty estimation by zero-shot inference with the VLM using multiple augmented views of an input. Then, by leveraging the feature representation ability of VLM, we obtain feature-based pseudo-labels via sample clustering in the feature space.
Prompt-feature consensus is introduced to select 
reliable samples based on the consensus between the two types of pseudo-labels.
We further propose
High-confidence Cross Supervision by to learn from samples with reliable pseudo-labels 
and the remaining unlabeled samples.
Additionally, we present an innovative open-set prompting 
strategy that filters irrelevant patches from whole slides
to enhance the quality of selected patches.
Experimental results on five public pathological image datasets for patch-level and slide-level classification showed that our method 
substantially outperformed zero-shot classification by VLMs, and was superior to existing 
noisy label learning methods.
The  code is publicly available at \url{https://github.com/HiLab-git/VLM-CPL}.
\end{abstract}

\begin{IEEEkeywords}
Pathological image classification, foundation model, pseudo-label, noisy label learning.
\end{IEEEkeywords}

\section{Introduction}
\label{sec:introduction}
Pathology image classification 
plays a crucial role in accurate cancer diagnosis, outcome prediction and treatment decision making~\cite{skrede2020deep}. 
Due to manual inspection for determining the characteristics of tumor's microenvironment is time-consuming,
automated recognition of diverse tissue types and subtypes 
within Whole Slide Images (WSIs) is highly desirable~\cite{skrede2020deep,campanella2019clinical}.
In recent years, 
Deep neural networks, notably Convolutional Neural Networks (CNNs) and Vision 
Transformers (ViTs), have boosted the accuracy and efficiency of analyzing 
microscopic pathology images~\cite{kather2016multi,abmil,shao2021transmil}. 
However, current 
advancements in digital pathology depend on large datasets annotated by experts, a process that is both labor-intensive 
and challenging to scale due to the vast size and complexity of pathology images, limiting their application in a wide 
range of pathological image analysis tasks.
Therefore, enhancing classification accuracy with a minimal annotation requirement, or ideally eliminating human 
annotations, has garnered significant interest within the digital pathology field.

Recently, some label-efficient techniques such as
Semi-Supervised Learning (SSL)~\cite{cps,luo2022urpc} 
and Active Learning~
(AL)~\cite{beluch2018power} 
have achieved promising results with reduced annotation cost. 
SSL methods typically utilize a small amount of labeled data along with a large set of unlabeled data, leveraging 
consistency regularization on unlabeled data or pseudo-labels to enhance the model's performance.
AL involves selectively querying the most informative samples 
from unlabeled data for human annotation, thereby improving 
model performance.
Despite these training paradigms can efficiently train 
high-performance models,
they still require a considerable amount of workload for annotators.

In recent years, large pre-trained Vision-Language Models (VLMs)~\cite{clip,plip,biomedclip,conch} have shown powerful 
zero-shot inference abilities for downstream classification tasks.
For example, Contrastive Language-Image 
Pre-training (CLIP)~\cite{clip} stands as a pioneering model distinguished by its utilization of image-text pairs and 
contrastive learning during network training. 
Utilizing CLIP-based models to obtain 
zero-shot pseudo-labels as a source of supervision is an intuitive approach for training a downstream 
network~\cite{enhancingclip},
which offers the possibility of completely getting rid of human annotations. For instance, 
Menghini et al.~\cite{enhancingclip} proposed to enhance CLIP~\cite{clip} by training with pseudo-labels iteratively 
via prompt tuning~\cite{vpt}. 
However, this work only selects pseudo-labels based on confidence, which still 
results in a large amount of noise in the selected samples. The iterative prompt tuning also increases the time 
consumption for downstream tasks. In addition, 
that method focuses on patch-level classification, and cannot be directly applied to WSI classification.

To address these issues, we propose a novel method Consensus pseudo-labels from VLM (VLM-CPL) for human 
annotation-free pathological image classification, which
can train high-performance classifiers effectively and efficiently with the help of VLMs, and can be applied to both patch-level and WSI-level classification tasks.
Firstly, unlike existing methods~\cite{luo2022urpc,OEEM,cps} 
that rely on a small set of labeled images or weak annotations to generate pseudo-labels, we employ a 
pre-trained VLM for zero-shot inference based on prompt on the training set to obtain pseudo-labels.
Secondly, considering the low accuracy of VLM's zero-shot 
inference on downstream datasets with domain shift, we additionally leverage the strong feature representation ability of VLM to obtain another type of pseudo-labels via clustering in the feature space.
A novel module named Prompt-Feature Consensus (PFC) is proposed to select 
reliable samples by considering consensus between the two types of pseudo-labels.
Finally, VLM-CPL uses High-confidence Cross 
Supervision (HCS) to learn from the selected samples with reliable pseudo-labels and the remaining unlabeled ones.
Our major contributions are:
\begin{itemize}
    \item A novel framework VLM-CPL is proposed for human annotation-free pathological image classification by leveraging the pre-trained VLMs.
    \item Two selection strategies are proposed to select high-quality pseudo-labels for model training. First, Multi-View Consensus (MVC) is based on multiple 
    random augmentations to identify confident predictions.
    Second, Prompt-Feature Consensus (PFC) is introduced to select 
    reliable samples by considering consensus between the prompt-based and feature-based pseudo-labels.
    \item To leverage samples with reliable pseudo-labels and other unlabeled samples, a High-confidence Cross Supervision (HCS) strategy is proposed for patch classification.
    \item {To deal with WSIs where patch-level and slide-level class label may mismatch, we propose an Open-Set Prompting (OSP) method by considering non-target classes to select reliable patches for WSI classification.
}
\end{itemize}
We conducted experiments on five public datasets, i.e., three patch-level and two WSI-level datasets 
spanning tissues from the colon, lung, prostate, and kidney, to verify the effectiveness of our method. 
The experimental results demonstrated that VLM-CPL exhibits superior performance across all five datasets, achieving an 
average accuracy improvement of 18.8\% without any human annotation compared with direct using VLMs for zero-shot 
inference.

\vspace{-1mm}
\section{Related works}

\subsection{Pathological Image Classification}
Pathological image classification techniques can be roughly divided into patch-level and WSI-level classification methods.
For patch-level pathological image classification, the prevalent strategy is to train CNNs or ViTs with fully supervised learning, and most works concentrate on network architecture and loss function design to enhance classification accuracy.
Lin et al.~\cite{lin2022pdbl} proposed a lightweight plug-and-play module to
construct a multi-resolution image pyramid for each patch to improve classification accuracy.
Moyes et al.~\cite{moyes2023multi} developed a Multi-Channel Auto-Encoder for robust feature representations against scanner-induced appearance variations.
Xue et al.~\cite{xue2021selective} utilized Generative Adversarial Networks (GAN) to synthesize histopathological patches, thereby enhancing feature representation and boosting classification accuracy.

As a WSI has a very high resolution (up to 100,000$\times$100,000), 
it can only be treated as a bag of multiple instances 
(patches), where only the bag-level label is given, with 
instance-level labels unknown. Thus, classification of WSIs 
is a Multiple Instance Learning (MIL) problem~\cite{abmil,CLAM,shao2021transmil,mambamil}. 
The essence of MIL lies in aggregating predictions or features 
from multiple instances to obtain slide-level results.
For example, ABMIL~\cite{abmil} derives attention scores from instance representations, with the scores indicating the significance of the respective patches.
CLAM~\cite{CLAM} incorporates an auxiliary task within the MIL framework to assess the relevance of instances based on attention size, and 
TransMIL~\cite{shao2021transmil} uses self-attention to capture patch relationships for WSI classification.

However, these methods are developed for fully supervised 
learning, restricting their applicability in scenarios lacking annotated 
training images. In contrast, our work aims to 
train a pathological image classification model without any human 
annotations by leveraging VLMs to generate pseudo-labels, and it 
can be applied to both patch-level and WSI-level classification 
tasks. 


\begin{figure*}[t] \centering
\includegraphics[width=\textwidth]{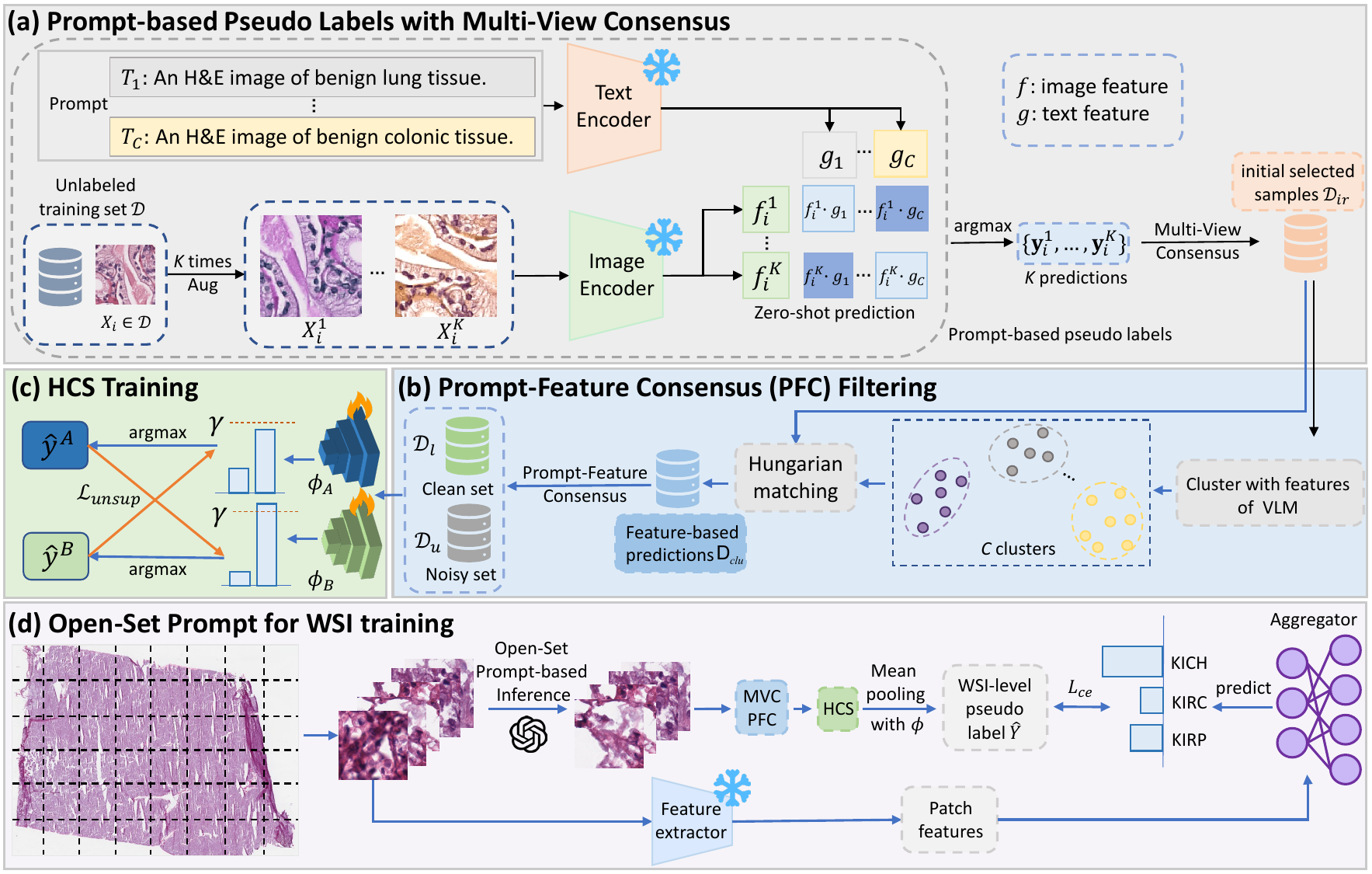}
\caption{{Overall framework of VLM-CPL.} For unlabeled training images, we first obtain pseudo-labels using prompt-
based inference of a VLM that are filtered by multi-view consensus (a). They are further filtered by considering the 
consensus with another type of pseudo-label using feature clustering (b). Then, High-confidence Cross Supervision 
(HCS) is proposed to train the patch classifier from filtered samples with reliable pseudo-labels and the remaining 
unlabeled ones (c).
For WSI classification tasks where patch-level and 
WSI-level label sets may mismatch (d), we further propose open-set prompt to filter  patches that are irrelevant to the target classes. Mean pooling of patches is employed to obtain WSI-level pseudo-labels that are used to train a learnable aggregator for WSI-level prediction.
{Note that the text encoder, image encoder and feature extractor are frozen, while only the classifiers $\phi_{a}$ and $\phi_{b}$ are trainable.}} \label{overall}
\end{figure*}

\subsection{Vision-Language Model}
Recently, large pre-trained VLMs have shown great feature representation and zero-shot inference capabilities. 
For example, the Contrastive Language-Image Pre-training (CLIP)~\cite{clip}  has showcased strong 
zero-shot inference capabilities attributed to its comprehensive training dataset of 400 million image-text pairs, 
significantly enhancing its ability to generalize across various tasks without training 
examples.
Similar to CLIP~\cite{clip}, ALIGN~\cite{align} was pre-trained on over 100 million noisy image-text pairs using contrastive learning.
In the medical imaging community, there have been several works that follow the CLIP~\cite{clip} approach.
For instance, 
BioMedCLIP~\cite{biomedclip} leveraged 15M
figure-caption pairs extracted from PubMed for training.
MI-Zero~\cite{mizero} was pre-trained on over 33k
histopathology image-caption pairs, and  
PLIP~\cite{plip} was trained on over 200k pathological image and description pairs derived from the Twitter platform. 
Similarly, CONCH~\cite{conch} and Quilt-1M~\cite{quilt-1M} were trained on a dataset exceeding 1 million 
image-caption pairs using a contrastive loss for optimization.
However, these models may have dropped performance due to the domain shift between the pre-training and downstream datasets. Adapting these models to downstream tasks without human labeling is an urgent issue that needs to be addressed in clinical applications.

\subsection{Noisy Label Learning}
To deal with noisy labels, several works tried to select clean samples
for better training the model. For instance,
Co-teaching~\cite{han2018co-teach} simultaneously trains two networks, with each network selecting samples based on low 
loss values from the other for training. Co-teaching+~\cite{yu2019co-teaching+} enhances performance by integrating the 
``Update by Disagreement'' strategy with the original Co-teaching approach~\cite{han2018co-teach}.
DivideMix~\cite{dividemix} utilizes a Gaussian mixture model to differentiate between clean and noisy labels based on 
the distribution of loss values, where small losses are indicative of clean labels and larger losses signify noisy 
labels.
HAMIL~\cite{zhong2023hamil} employs two networks to supervise a third one, leveraging knowledge 
distillation to mitigate the impact of noise.
Besides, other methods focus on novel loss functions or training frameworks.
Zhang et al.~\cite{GCE} proposed a noise-robust generalized cross entropy loss that is
a generalization of mean absolute error loss and categorical cross entropy.
Liu et al.~\cite{ELR} found early learning and memorization during training with noisy labels, and introduced a regularization term to prevent the direct memorization of noisy labels.
{However, these methods typically rely solely on the label information during training, neglecting the feature space of the model for generating pseudo-labels or ignoring inter-sample similarity, thereby limiting their performance. }

\section{Methods}
Let $\mathcal{D} =\{X_i\}_{i=1}^N$ denote an unlabeled training set with $C$ classes, and
$N$ is the sample number.
This paper aims to train a classifier from $\mathcal{D}$  without human-provided annotations. As illustrated in 
Fig.~\ref{overall}, the proposed VLM-CPL contains three stages for patch-level 
classification: 
1) Prompt-based pseudo-labels with Multi-View Consensus (MVC);
2) Feature-based pseudo-labels and prompt-feature consensus-based filtering, leading to two complementary subsets: 
$D_l$ with clean pseudo-labels and $D_u$ with unannotated samples;
3) High-confidence Cross Supervision (HCS) to train the classifier using $D_l$ and $D_u$.
For WSI classification tasks, we propose open-set prompt to obtain WSI-level pseudo-labels from patch-level labels, and train a patch aggregator to obtain the prediction for a WSI, as shown in Fig.~\ref{overall}(d).

\subsection{Prompt-based pseudo-labels with Multi-View Consensus}
\label{sec3.1}
{
Recent Vision-Language Models (VLMs)~\cite{clip,plip,biomedclip} with
zero-shot inference abilities, are built on a CLIP-like 
architecture~\cite{clip} that consists of an image encoder \( 
E_{img} \) and a text prompt encoder \( E_{text} \) to align visual and 
textual features in a shared feature space.}
{For a given image $X_i$ (i.e., a  patch in this section), its image feature representation 
is $f_i= E_{img}(X_i)$.
For a downstream classification task with $C$ classes, the text prompt for the $c$-th class is $T_c = \text{``An H\&E image of \{CLS$_c$\}}$," where \texttt{CLS$_c$} represents the name of the $c$-th class. The corresponding text feature is  \( g_c = E_{text}(T_c) \). 
Let \( p_i \in [0, 1]^C \) denote the probability vector for classifying image \( X_i \). The \( c \)-th element of \( p_i^c \), representing the probability of \( X_i \) belonging to class \( c \), is calculated by:} 
\begin{equation}
    p_i^c = \frac{e^{sim(f_i, g_c)/\tau}}{\sum_c e^{sim(f_i, g_c)/\tau}}
    \label{eq1}
\end{equation}
where $sim(\cdot,\cdot)$ and $\tau$ denote the cosine similarity and the temperature, respectively.
{We denote the pseudo-label for $X_i$ as $\hat{Y}_i=\text{argmax}(p_i)$, and the training set is represented as $\mathcal{D}_p = \{(X_i, \hat{Y}_i)\}_{i=1}^N$.}
Note that $\mathcal{D}_p$ contains a large amount of noise 
due to the domain gap between the pre-training and target 
datasets, directly training a network from  
$\mathcal{D}_p$ using standard supervised learning may 
lead to model collapse.


\subsubsection{Multi-View Consensus (MVC)}
To deal with noisy labels, based on the assumption that uncertainty information can 
effectively indicate the quality of pseudo-labels~\cite{uncertainty-view,luo2022urpc, wang2019aleatoric,gaillochet2022taal}, 
we first introduce MVC based on multiple 
random augmentations to select confident predictions.
{It is inspired by Test-Time Augmentation (TTA)~\cite{wang_tta} that generates diverse views of the input through various transformations, allowing the model to capture variations naturally occurring during data acquisition~\cite{ayhan2018test} to estimate the aleatoric uncertainty on the test sample~\cite{wang_tta}.}
{
Let $\mathcal{T}$ represent a set of data augmentation operations, which contains two types: spatial transforms (e.g., random crop, rotation, flipping) and color transforms (e.g., ColorJitter).}
{Keeping the text prompt unchanged, we generate \( K \) randomly augmented versions of \( X_i \) and send them into the VLM model as described in Eq.~\ref{eq1}. 
The prediction for the \( k \)-th augmented version is denoted as \( \hat{Y}^{(k)}_i \).  The average prediction is denoted as $\bar{p}_i = (\sum_k 
\hat{Y}^{(k)}_i)/K$, and the uncertainty is estimated as 
    $v_i = -\sum_{c=0}^{C-1}\bar{p}^c_i\log \bar{p}^c_i$.}
A lower $v_i$ indicates a stronger consensus between the 
$K$ predictions under augmentation, and thus $\hat{Y}_i$ 
is more reliable. 
{Let \( v_M \) represent the \( M \)-th percentile of \( v_i \) across the entire dataset. It is used as a threshold to select an initial reliable subset of \( \mathcal{D}_p \). The resulting subset is denoted as \( \mathcal{D}_{ir} \).}
\begin{equation} 
    \label{eq:UF}
    \mathcal{D}_{ir} = \{(X_i,\hat{Y}_i)~|~X_i \in 
    \mathcal{D}_p, ~v_i \leq v_M\}
\end{equation}
where the size of $\mathcal{D}_{ir}$ is 
$N^{'}=N\times M\%$.
By leveraging MVC to identify potentially noisy samples,
the subset $\mathcal{D}_{ir}$
contains fewer low-confidence samples, which ensures high-quality pseudo-labels are obtained in $\mathcal{D}_{ir}$.

\subsubsection{Class-aware Multi-View Consensus (CMVC)}{
In many real-world pathology image datasets, class imbalance 
is a significant challenge. When generating pseudo-labels based on MVC that selects the top $M$\% confident samples from the entire dataset, class imbalance may be introduced due to the potential bias of the VLM to some easy classes, where hard classes with higher uncertainty may be rejected. 
To alleviate the class imbalance problem, we introduce a variant of MVC, i.e., CMVC that selects the top $M$\% confident samples for each 
class based on their pseudo-labels. 
This ensures that the selected samples in $\mathcal{D}_{ir}$ contain all the classes, and make the distribution of 
pseudo-labels does not favor a particular class or a few dominant classes.}

\subsection{Prompt-Feature Consensus Filtering}
\label{sec3.2}
Pre-trained VLMs can not only perform zero-shot inference but also 
obtain powerful image feature representations~\cite{clip,plip,conch}. 
{In addition to obtaining pseudo-labels through prompt-based inference (as described in Eq.~\ref{eq1}), the inter-sample similarity in the feature space can also be utilized to enhance the selection process.}
{Since both methods utilize the same image encoder to obtain pseudo-labels, samples with inconsistent pseudo-labels are more likely to be unreliable~\cite{wu2022mutual,gaillochet2022taal}.}
{Therefore, we propose a Prompt-Feature Consensus (PFC) filtering method to further obtain more reliable pseudo-labels from \( \mathcal{D}_{ir} \).}

Firstly, we use K-means++~\cite{kmeans++} to cluster samples in $D_{ir}$ with a cluster number of $C$, using features 
extracted from $E_{img}$ of the VLM.
We denote the clustering results as
$\{(X_i, O_i)\}_{i=1}^{N{'}}$, where $O_i \in \{0, 1, ...., C-1\}$  
is the cluster label of $X_i$.
{Since \( O_i \) represents a cluster label derived from the clustering process, it does not directly align with the predefined class labels. For example, cluster 1 from the clustering process might represent normal tissue, while the predefined class labels designate 0 for normal tissue and 1 for cancer tissue. To address this mismatch, we use a bijection function \( h(\cdot) \) to map \( O_i \) to the corresponding class label, resulting in a cluster-based pseudo-label \( \Tilde{Y}_i = h(O_i) \). The mapping function \( h(\cdot) \) is computed using Hungarian matching~\cite{zhu2011group}, which optimizes the alignment by maximizing the consensus between the cluster-based pseudo-labels and the labels in \( \mathcal{D}_{ir} \).}
\begin{equation}
    \underset{h}{\mathrm{argmax}} \sum_{i=1}^{N^{'}} \mathbbm{1}[\hat{Y}_i==h(O_i)]
\end{equation}
where $\mathbbm{1}[\cdot]$ is a binary indicator.
After solving $h(\cdot)$, we denote the dataset with cluster-based pseudo-labels as $\mathcal{D}_{clu}=\{(X_i, \Tilde{Y}_i)\}^{N{'}}_{i=1}$, where $\Tilde{Y}_i=h(O_i)$.
$\mathcal{D}_{ir}$ and $\mathcal{D}_{clu}$ contain the prompt-based pseudo-labels and feature similarity-based pseudo 
labels, respectively.
{By taking the intersection of these two results, we can filter out inconsistent pseudo-labels, which are more likely to be noisy, and retain only the reliable ones. Specifically, we select samples with consistent labels between \( \mathcal{D}_{ir} \) and \( \mathcal{D}_{clu} \), resulting in the final filtered subset \( \mathcal{D}_l \):}
\begin{equation}
   \mathcal{D}_l = \{(X_i, \hat{Y}_i)~|~X_i \in \mathcal{D}_{ir}, ~\hat{Y}_i =\Tilde{Y}_i\}
\end{equation}
$\mathcal{D}_l$ can be considered as a clean subset, and we abandon the $\hat{Y}_i$ for other samples due to their low reliability, and denote them as an unannotated subset $\mathcal{D}_u =\mathcal{D}-\mathcal{D}_l$.

\subsection{High-confidence Cross Supervision}
\label{sec3.3}
After filtering the reliable prompt-based pseudo-labels with PFC, 
as the clean subset $\mathcal{D}_l$ has a smaller size than the original dataset $\mathcal{D}$, only using $\mathcal{D}_l$ to train a downstream model may limit the performance. To better leverage all the samples in $\mathcal{D}$, we propose 
High-confidence Cross Supervision (HCS) that takes a combination of $\mathcal{D}_l$ and $\mathcal{D}_u$ to train the downstream model. 
HCS is inspired by CPS~\cite{cps} that utilizes 
two networks to generate pseudo-labels of unlabeled images for each other. 
CPS does not 
consider the quality of pseudo-labels, which may hinder the performance. To address this problem, HCS selects
high-confidence pseudo 
labels for training.

Specifically, let $\phi_A$ and $\phi_B$ denote two parallel networks with the same architecture for the downstream patch-level classification task.
A patch-level image $X_i$ is randomly augmented 
into two views 
that are sent to $\phi_A$ and $\phi_B$ respectively, and the outputs are denoted as $p_i^A$ and $p_i^B$, respectively.
For samples in  $\mathcal{D}_u$, we convert $p^A_i$ and $p^B_i$ into one-hot pseudo-labels
$\hat{y}^A_i$ and $\hat{y}^B_i$ by argmax operation respectively.  
To filter out low-quality pseudo-labels, a confidence threshold $\gamma$ is adopted, and the unsupervised loss is defined as:
\begin{equation} \label{eq5}
  \begin{split}
  \mathcal{L}_{unsup}^A &= 
  \mathbbm{1}[max(p_i^B)>\gamma] \cdot L_{ce}(p_i^A, \hat{y}_i^B)    \\
  \mathcal{L}_{unsup}^B &= 
  \mathbbm{1}[max(p_i^A)>\gamma] \cdot L_{ce}(p_i^B, \hat{y}_i^A) \\
  \mathcal{L}_{unsup} &= (\mathcal{L}_{unsup}^A+\mathcal{L}_{unsup}^B)/2
  \end{split}
\end{equation}
where $max(p_i^A)$ denotes the maximal probability value across all the classes in $p_i^A$.
$L_{ce}$ denotes the cross-entropy loss.
{For the clean subset $\mathcal{D}_l$, we use a pseudo-label loss $\mathcal{L}_{pl}$ implemented by $L_{ce}$: $\mathcal{L}_{pl}=(L_{ce}(p_i^A, \hat{Y})+L_{ce}(p_i^B, \hat{Y}))/2$.}
Finally, the overall loss is $\mathcal{L}=\mathcal{L}_{pl}+\lambda \mathcal{L}_{unsup}$,
where $\lambda$ is the weight of the unsupervised loss.

\begin{figure}[t] \centering
\includegraphics[width=0.48\textwidth]{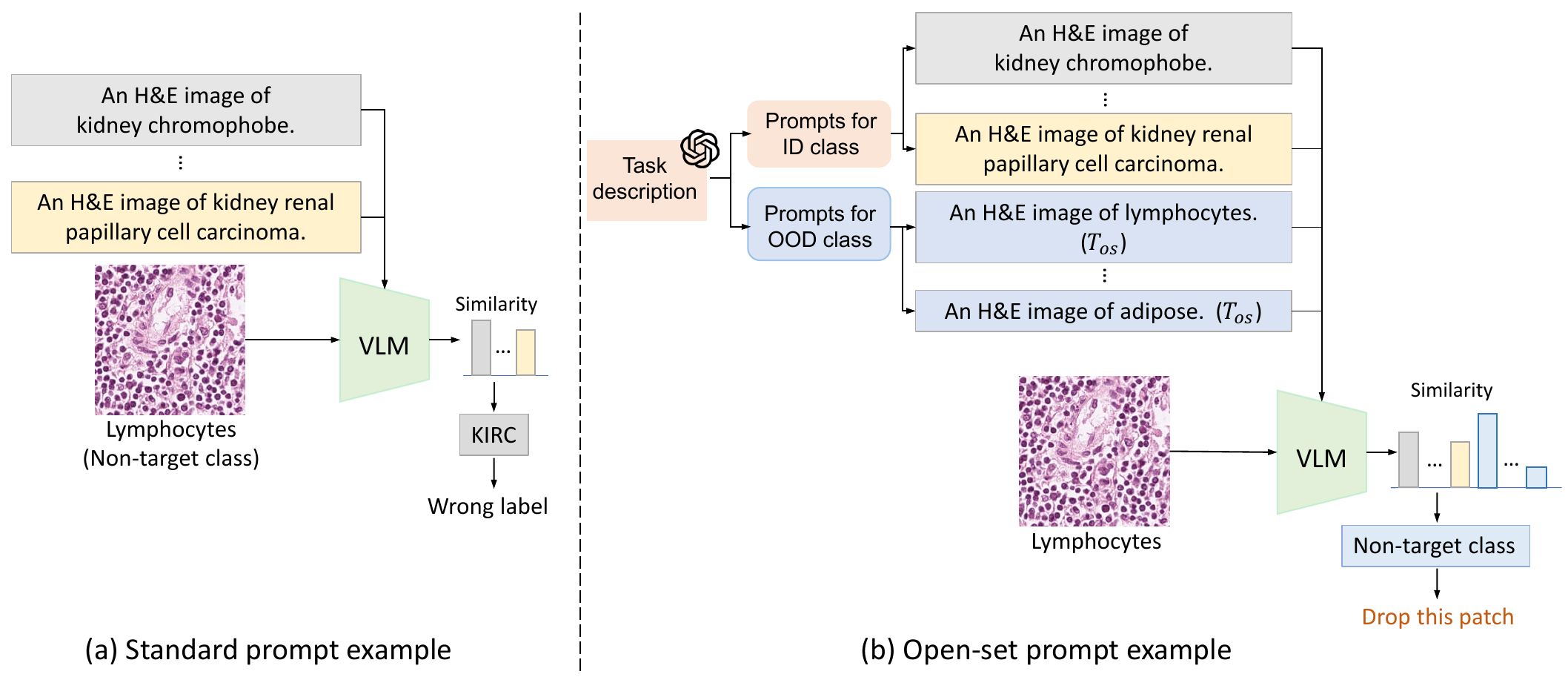}
\caption{\textcolor{black}{Comparison between standard prompt (a) and our proposed Open-Set Prompt (OSP) (b). The former obtains incorrect labels for samples of non-target classes, while the latter is able to reject patches that are irrelevant to the target WSI classes, which ensures the quality of pseudo-labels for selected samples.}}
\label{ood}
\end{figure}
\subsection{Extension from Patch-level to WSI-level Classification}
The above MVC, PFC and HCS are designed for patch-level classification tasks, and 
$\phi_A$ or $\phi_B$ trained with these modules cannot be directly applied to WSI classification tasks due to two main issues: First, WSI-level labels may have a mismatch with patch-level labels. For example, 
for a kidney tumor WSI classification task, the candidate labels for WSIs are Kidney Chromophobe Renal Cell Carcinoma 
(KICH), Kidney Renal Clear Cell Carcinoma (KIRC), and Kidney Renal Papillary Cell Carcinoma (KIRP). However, the patches in a kidney tumor WSI may belong to none of these classes, such as lymphocytes and adipose~\cite{qu2023openal}. As VLMs can only take patches for zero shot inference, using the prompts with class labels of \{KICH, KIRC, KIRP\} will force the VLM to classify a patch into one of these closed-set labels, leading to wrong pseudo-labels for most patches that are non-tumor, as illustrated in Fig.~\ref{ood}(a). Second, after patch-level predictions are obtained by $\phi_A$ or $\phi_B$, they should be aggregated into a WSI-level label. Though some simple methods such as mean pooling can achieve this goal, the performance may be limited, and a learnable aggregator is desired for better performance. 


To address these issues, we further propose two modules that extend our method for WSI classification tasks: 1) an Open-Set Prompting (OSP) method that avoids incorrectly classifying an irrelevant patch into one of the WSI-level labels for pseudo-label generation; 2) Training an additional WSI classifier that aggregates patch-level labels into WSI labels, as shown in Fig.~\ref{overall}(d).

\subsubsection{Open-set prompt}
The OSP strategy is illustrated in Fig.~\ref{ood}(b).
Let $W_j$ denote the $j$-th WSI in the training set, and is divided into patches via sliding window with non-tissue background patches dropped, leading to a set of $S_j$ patches $W_j=\{X_i\}^{S_j}_{i=1}$. Note that $S_j$ is not a constant, but instead varies in different WSIs. The previous symbol $\mathcal{D}$ denoting the training set is therefore the union of all $W_j$ in a WSI classification task. 
\textcolor{black}{To avoid classifying a non-tumor patch into one of the tumor classes (WSI labels) as mentioned above, we introduce an extra prompt $T_{os}$ that does not correspond to any of the $C$ target WSI-level classes, but is relevant to other types of patches that may appear in WSIs. To construct $T_{os}$, we query a large language model (GPT-4o) with a brief description of the classification task and the list of in-distribution categories, asking it to suggest plausible out-of-distribution tissue types. This leverages the broad domain knowledge encoded in the GPT-4o and enables generalization to other tasks without requiring expert involvement. For instance, the prompt ``\texttt{An H\&E image of lymphocytes}'' is one such GPT-generated example.}
Assume there are $Q$ non-target classes, we introduce a new class index $c' \in \{0, 1, 2, ..., C-1, C, ..., C+Q-1\}$, 
where $c'>=C$ means the non-target class.
Similar to Section~\ref{sec3.1}, we use Eq.~\ref{eq1} to compute the probability $p_{c^{'}}$ of class $c'$.
The selection rule for target class-relevant instance bag $W_j^{'}$ is formulated as:
\begin{equation}
    W_j^{'} = \{(X_i, \hat{Y}_i)~|~\hat{Y}_i < C, X_i \in W_j\}
\end{equation}
where $\hat{Y}_i$ is the pseudo-label of $X_i$ based on argmax of $p_{c^{'}}$.

\begin{table*}[t]\footnotesize
    \centering
    \caption{Information of five downstream public datasets of pathological images used in our experiments.}
    \begin{tabular}{l|ccccc}
    \hline
    \multicolumn{1}{c|}{} & Sample type & Train / Test  & Image size & Organ          & Class number \\ \hline
    HPH~\cite{salvi2021hybrid}   & patch &  17,126 / 8,177         & 256$\times$256               & prostate       & 2         \\ 
    LC25K~\cite{LC25K}  & patch & 20,000 / 5,000         & 768$\times$768               & lung and colon & 5         \\ 
    NCT-CRC-HE-100K~\cite{kather2016multi}  & patch &  71,547 / 17,887   & 224$\times$224   & colon   & 9   \\ \hline
    DigestPath~\cite{digestpath} & WSI &   528 / 132           & $\sim$5,000$\times$5,000       & colon          & 2         \\ 
    TCGA-RCC  & WSI & 249 / 107    & $\sim$50,000$\times$35,000   & kidney   & 3     \\ \hline
    \end{tabular}
    \label{tab:dataset}
\end{table*}
For an unannotated WSI classification dataset, we apply OSP to MVC described in Section~\ref{sec3.1} to obtain tumor-relevant patches as the initial reliable subset $\mathcal{D}_p$ that is the union of all $W'_j$. Then we apply PFC to $\mathcal{D}_p$ to obtain $\mathcal{D}_l$ and then train patch-level classifiers $\phi_{A}$ and $\phi_{B}$ by taking $\mathcal{D} - \mathcal{D}_l$ as $\mathcal{D}_u$, following Section~\ref{sec3.2} and \ref{sec3.3}, respectively. 

\subsubsection{WSI-level training and inference}
With the trained patch-level classifier $\phi_A$ and $\phi_B$, we first use them to obtain patch-level prediction
$p_i=(\phi_A(X_i)+\phi_B(X_i))/2$.
The slide-level prediction scores are obtained by passing the $S_j$ patches in $W_j$
to a pooling operator such as average pooling~\cite{mizero}:
\begin{equation}
  \label{eq.mean}
  {{p}_{avg}} = \frac{1}{S_j} \sum_{i=1}^{S_j}p_i
\end{equation}
{where ${p}_{avg} \in \mathbb{R}^C $ represents
average prediction vector across all the patches.
The \(\text{argmax}\) operation is then used to ${p}_{avg}$ to convert it into a one-hot pseudo-label $\hat{Y}$.}
With the slide level pseudo-labels, we treat the WSI classification 
task as a MIL problem. 
Specifically, the cropped patches are fed 
into a feature extractor $\psi$ to obtain visual 
features, and they are aggregated by a learnable aggregator that 
is trained with the WSI-level pseudo-labels. In this work, the 
aggregator is implemented by CLAM~\cite{CLAM} based on an 
attention network, due to CLAM's robust performance in the 
literature~\cite{CLAM}.
After training the aggregator, it is applied to a testing WSI 
with features extracted by $\psi$ to obtain the WSI-level 
prediction.

\section{Experiments}
\subsection{Datasets}

We conducted experiments on five public pathological image datasets, as listed in Table~\ref{tab:dataset}. 
First, we used three datasets for patch-level classification:
1) \textbf{Human Prostate Histology 
(HPH)}~\cite{salvi2021hybrid}
dataset\footnote{https://data.mendeley.com/datasets/h8bdwrtnr5/1}
that comprises 738 
pathological images containing malignancies from 150 
patients. 
They were captured at a magnification of $\times100$ 
(0.934$\mu m$/pixel) with a size of 
1500$\times$1500 pixels. 
We randomly selected 500 images from 100 patients for 
training and used the remaining 238 images from 50 
patients for testing.
The original images were cropped patches with of 256$\times$256 pixels without overlap, leading 
to  17,126 patches (12,082 for cancer) in the training set 
and 8,177 patches (6,071 for cancer) in the testing set, 
respectively.
The patches were used for binary classification between normal and cancer.
2) 
\textbf{Lung and Colon histopathological image 
(LC25K)}~\cite{LC25K} 
dataset\footnote{https://huggingface.co/datasets/1aurent/LC25000} 
with 25,000 
patches and five classes: 
benign lung tissue {(L-NORM)}, 
lung adenocarcinoma {(L-TUM)}, 
lung squamous cell carcinoma {(L-SCC)}, benign colonic tissue {(C-NORM)}, 
and colon adenocarcinoma {(C-TUM)}. The patch size is 768$\times$768 pixels.
We randomly partition the dataset into a training set and 
a testing set with a ratio of 4:1. 3) 
\textbf{NCT-CRC-HE-100K}~\cite{kather2016multi} 
dataset\footnote{https://zenodo.org/records/1214456} that contains 100,000 patches of 
colorectal cancer pathology images with nine
fine-grained classes: 
adipose ({ADI}, 10\%), background ({BACK}, 11\%), debris
({DEB}, 11\%), lymphocytes ({LYM}, 12\%), mucus ({MUC}, 9\%), 
smooth muscle (MUS, 14\%), normal colon mucosa ({NORM}, 9\%), 
cancer-associated stroma ({STR}, 10\%),
and colorectal adenocarcinoma epithelium ({TUM}, 14\%). The resolution is 224$\times$224 pixels.
We excluded the background category, as it holds little significance for clinical 
diagnosis. The remaining patches were randomly split into 80\% for training and 20\% for 
testing.

We also used two public datasets for WSI-level classification: 1)
\textbf{DigestPath}~\cite{digestpath}
dataset\footnote{https://digestpath2019.grand-challenge.org/}
for binary classification of malignant and benign colon WSIs that were from four medical institutions of 
$\times$20 magnification (0.475$\mu m$/pixel), 
with an average size of 5,000$\times$5,000.
It comprises a total of 660 samples, with 250 positive 
cases and 410 negatives.
We randomly split them into 80\% and 20\% for training and testing, respectively. 2) \textbf{TCGA-RCC} dataset\footnote{https://portal.gdc.cancer.gov/} for classification
of three subtypes of kidney tumor. It contains a total of 356 WSIs, including 120 for kidney chromophobe renal cell
carcinoma (KICH), 119 for kidney renal clear cell carcinoma 
(KIRC), and 117 for kidney renal papillary cell carcinoma 
(KIRP). The average size is 50,000$\times$35,000.
We randomly divided them into 249 for training and 107 for testing, respectively.


\begin{table}[t]
\centering
\caption{Ablation study on the HPH dataset for path-level classification. $N_s$ is the number of selected samples in the training set.
The model architecture
is ResNet50.}
\resizebox{0.49\textwidth}{!}{
\begin{tabular}{c|ccc|ccc}
\hline
    \multirow{2}*{Method}& \multicolumn{3}{c|}{pseudo-label quality}  & \multicolumn{3}{c}{Testing performance}  \\ \cline{2-7} 
        & $N_s$& ACC & F1 & ACC & F1 &  Recall \\ \hline
Baseline   & \textbf{17126} & 0.645 & 0.642 & 0.759 & 0.703 &  0.718 \\
+ MVC      & 5137 & \underline{0.904} & \underline{0.890} & 0.847 & 0.779 & 0.756 \\
\textcolor{black}{+ Entropy filter} & \textcolor{black}{5137} & \textcolor{black}{0.819} & \textcolor{black}{0.795} & \textcolor{black}{0.730} & \textcolor{black}{0.712} & \textcolor{black}{0.799} \\
+ PFC & \underline{11578} & 0.811 & 0.803 & 0.842 & 0.780 & 0.764 \\ 
+ MVC + PFC   & \multirow{1}*{4893} & \multirow{1}*{\textbf{0.926}} & \multirow{1}*{\textbf{0.912}} & 0.848 & 0.812 & 0.831 \\ \hline
+ MVC + PFC + CPS & 4893 & 0.926 & 0.912 & \underline{0.850} & \underline{0.819} & \textbf{0.849} \\
+ MVC + PFC + HCS & 4893 & 0.926 & 0.912 & \textbf{0.871} & \textbf{0.836} & \underline{0.845} \\ \hline
\end{tabular}}
\label{tab:ablation2}
\end{table}

\begin{figure*}[h]
    \centering
    \includegraphics[width=1\textwidth]{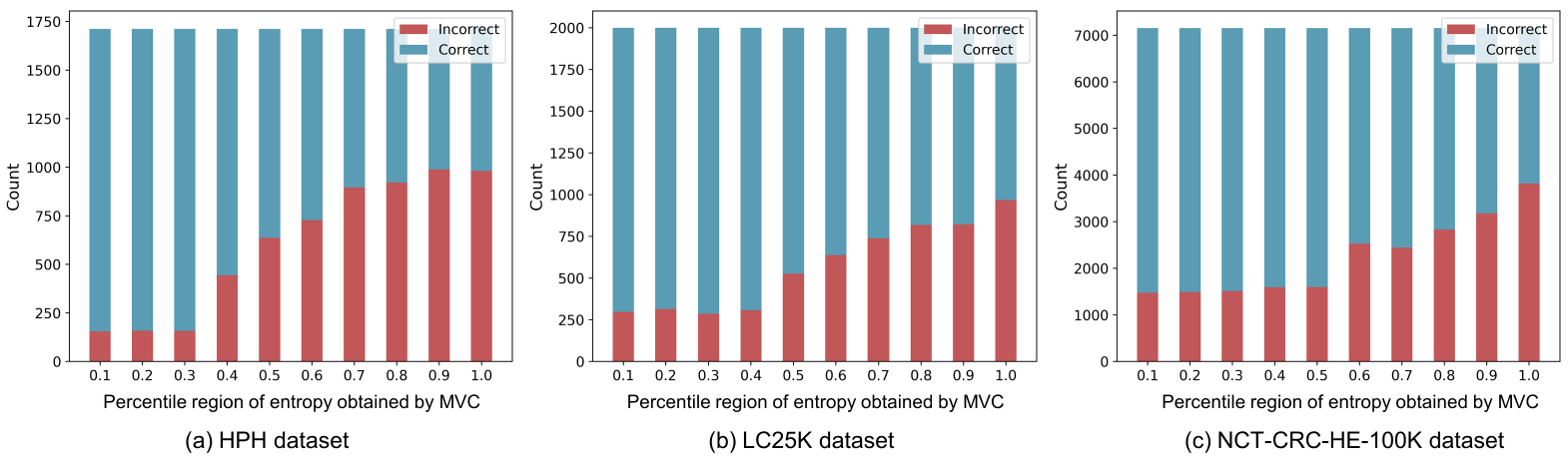}
    \caption{{Count of correct and incorrect pseudo-labels under different percentile intervals of entropy obtained by MVC. 
    The first bar shows the first 10\% samples with the lowest entropy, while the last bar shows the top 10\% samples with the highest entropy. }}
    \label{fig:MVC}
\end{figure*}
\subsection{Implementation Details}
The VLM-CPL framework was implemented in PyTorch,
and experimented with one NVIDIA GeForce RTX 3090 GPU. It is also deployed on the SenseCare platform~\cite{Sensecare2024} to support clinical research.
For pseudo-label generation, we considered two leading VLMs~\cite{plip,biomedclip} 
that are specially trained for 
pathological images: 1) PLIP~\cite{plip} that was trained on over 200k pathological 
image-text pairs from 
Twitter; 2) BioMedCLIP~\cite{biomedclip} that was trained on over 15M biomedical image-text pairs from 
PubMed. Note that they do not have overlap with the downstream datasets used in the experiment. PLIP was used for HPH and LC25K datasets, and ensemble of PLIP and BioMedCLIP was used for the other three datasets according to the best zero-shot inference performance.
The image encoder of PLIP and BioMedCLIP was a 
ViT-B/16~\cite{vit2021} with an output dimension of 768.

For training downstream patch-level classifiers $\phi_A$ and $\phi_B$, we 
employed the UNI encoder~\cite{UNI} 
and added a classification head that 
comprises two fully connected layers with 256 and $C$ output 
nodes respectively. 
Following~\cite{hu2021lora}, we added a Low Rank Adaptation (LoRA)~\cite{hu2021lora} 
layer for each 
Transformer layer for efficient fine-tuning with the SGD 
optimizer, a weight decay of $8\times10^{-4}$ and epoch number of 200. 
The learning rate was initialized to $10^{-4}$ and 
decayed by 0.1 every 100 epochs.
Following~\cite{plip}, we set $\tau=4.5871$, and based on ablation studies on the HPH dataset, we set $M=30$, $K=20$, $\gamma=0.8$ and $\lambda=1$.
For the LC25K and NCT-CRC-HE-100K datasets, 
the hyper-parameter setting was the same as that for HPH except that $K = 10$ and epoch = 300.
The batch size was 128 (64 labeled and 64 unlabeled images), and
random flipping, rotation and color jitters were used for data augmentation.

For training the aggregator for WSI classification on the DigestPath and TCGA-RCC datasets, we used the CLAM~\cite{CLAM} method, with UNI~\cite{UNI} as the feature extractor $\psi$.
We used the Adam optimizer with a 
learning rate of $2\times10^{-4}$ and a batch size of 1 WSI 
and 50 epochs.
The patch size was set to 256$\times$256 for computational feasibility. As the label set (benign and malignant) was 
the same for patches and WSIs on the DigestPath dataset, we only applied OSP to the TCGA-RCC dataset, and used $Q=2$ 
non-target classes: lymphocytes and adipose.
For both patch-level and WSI-level classification tasks, evaluation metrics include Accuracy (ACC), F1 
score and Recall.
For multi-class classification, the metrics were computed by macro-average.

\begin{figure*}[t]
    \begin{flushleft}
    \includegraphics[width=0.95\textwidth]{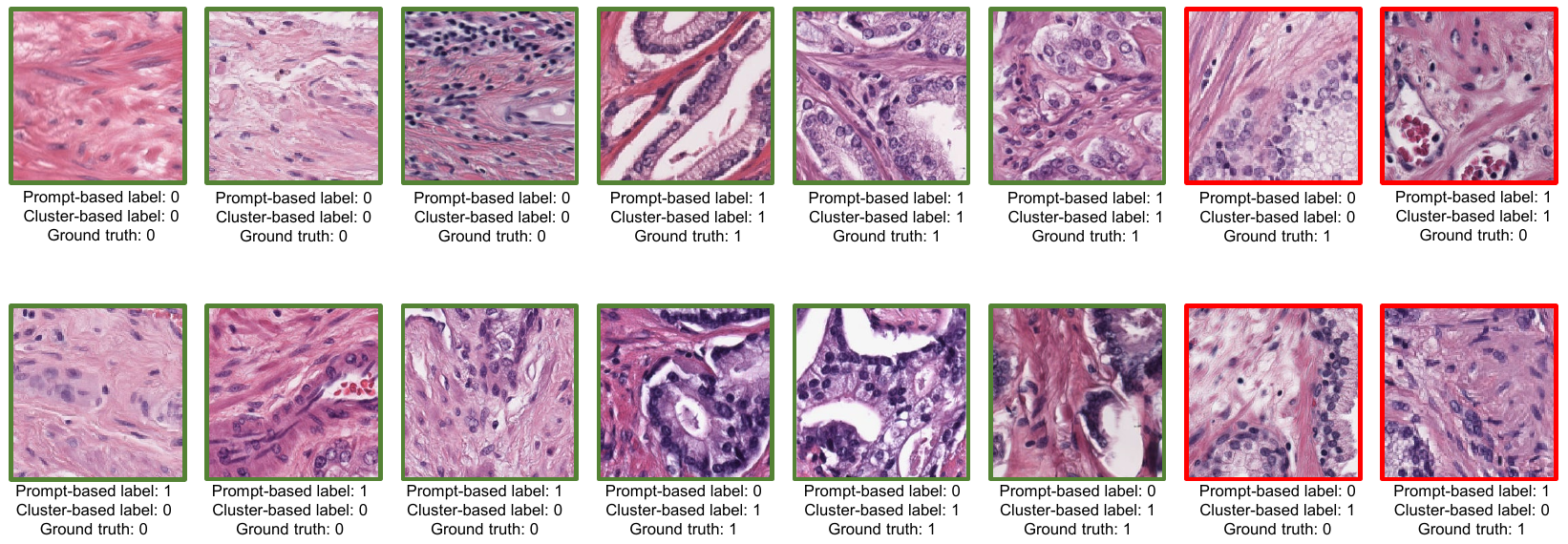}
    \centering
    \caption{{Examples illustrating how PFC works on the HPH dataset. Samples with consistent cluster-based and prompt-based labels are kept (first row).
    Samples with inconsistent pseudo-labels are filtered out (second row). Green and red boxes show success and failure cases, respectively.}}
    \label{fig:PFC}
    \end{flushleft}
\end{figure*}

\begin{figure*}
    \centering
    \includegraphics[width=0.95\textwidth]{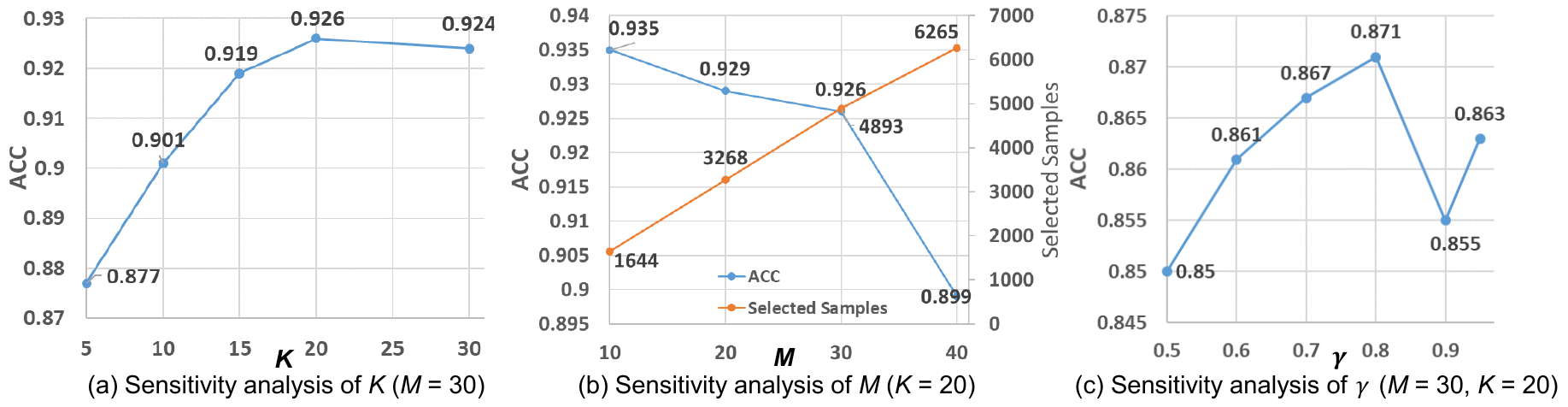}
    \caption{The impact of hyper-parameters on the HPH dataset. Note that $\gamma$=0.5 is the original CPS~\cite{cps}.}
    \label{fig:hyper}
\end{figure*}

\subsection{Results for Patch-level Classification}



\subsubsection{Ablation study}
For ablation study, we first evaluate the quality of pseudo 
labels on the training set, and set the baseline as directly 
using PLIP on the HPH training set for zero-shot inference.
Our MVC was compared with using entropy of $p_i$ in a 
single-forward pass for sample filtering (keeping the $M$\% most 
confident samples and $M$ = 30). 
We also compared the performance on the testing set for models 
trained with these pseudo-labels using ResNet50. 
Results in Table~\ref{tab:ablation2} show 
that the baseline method leads to the highest number of samples for training the downstream model, but the pseudo 
label has a low accuracy of 0.645. By using MVC to reject some low-quality pseudo-labels, the accuracy of remaining 
samples was 0.904.
\textcolor{black}{To demonstrate the superiority of MVC, we also compared it with 
a simple entropy filter, i.e., using the entropy through a single forward pass 
as the metric (keeping samples with the lowest entropy). Compared with our MVC, 
this approach yields pseudo-labels with 8.5 percentage points lower accuracy and 
9.5 percentage points lower F1 score. On the testing set, the accuracy and F1 
score decrease by 11.7 and 6.7 percentage points, respectively.}
Combining MVC and PFC further improved the accuracy of selected samples to 0.926.
Despite the reduction in the number of selected samples, the 
higher quality of the selected ones was beneficial for training, 
leading to an accuracy of 0.848 on the testing set, which 
outperformed the other sample selection strategies.
Then, using HCS for training further improved the accuracy on the testing set to 0.871.
Replacing HCS by CPS~\cite{cps} decreased the accuracy by 2.1
percentage points, demonstrating the importance of selecting 
confident samples during training with $\mathcal{D}_l$ and 
$\mathcal{D}_u$.

\begin{table}[t]
\centering
\caption{{Ablation Study of data augmentations on the HPH dataset.}}
\begin{tabular}{ccc|c|cc}
\hline
\multicolumn{3}{c|}{{Spatial}} & {Color} & {ACC} & {F1} \\ \hline
 {Crop}   &  {Rotation}  & {Flip} &   {ColorJitter}  &  {-} &  {-} \\ \hline
   &         &         &       &   {0.645} &  {0.642} \\
 {\checkmark}  &         &         &       &   {0.668} & {0.664} \\
&      {\checkmark}    &         &       &  {0.676}   & {0.675} \\
&         &      {\checkmark}    &       &   {0.662} & {0.661} \\
&         &         &    {\checkmark}    &   {0.867}  & {0.854} \\ \hline
{\checkmark}  &  {\checkmark}  &  {\checkmark}   &     & {0.684}  &  {0.682}\\
    {\checkmark}    &     {\checkmark}    &    {\checkmark}     & {\checkmark}  &  {\textbf{0.904}} & {\textbf{0.890}} \\ \hline
\end{tabular}
\label{tab_data_aug}
\end{table}
{Fig.~\ref{fig:MVC} shows the distribution of correct and 
incorrect pseudo-labels under different percentile regions of the entropy obtained 
by MVC, where a larger entropy value indicates higher inconsistency between $K$ predictions in MVC. It can be seen that there is a clear correlation between 
entropy values and the reliability of predictions. In the HPH dataset, as the entropy values increase, the count of incorrect predictions increases, indicating that higher
inconsistency is associated with lower reliability. This trend is also 
evident on the LC25K and NCT-CRC-HE-100K datasets, where high-inconsistency 
regions correspond to a significantly higher proportion of misclassifications. 
Consequently, these observations suggest that selecting samples with consistent 
predictions during MVC can lead to the acquisition of more reliable pseudo-labels.}

{Fig.~\ref{fig:PFC} presents examples from the HPH dataset to illustrate how PFC works.
The first and second rows show consistency and inconsistency between the prompt-based and cluster-based pseudo-labels, respectively.
In the first row, green boxes indicate success cases, as their pseudo-labels are correct and have been selected, while red boxes show failure cases, whose pseudo-labels are incorrect but are still selected by PFC. These failure cases often stem from challenging samples near the decision boundary, where zero-shot inference generates incorrect predictions that are still grouped into the correct cluster, causing PFC to fail in filtering these errors. However, PFC can reject most unreliable pseudo-labels, and improve the quality of samples after selection, as shown in Table~\ref{tab:ablation2}.}





\subsubsection{Data augmentations in MVC}
{
To investigate the impact of different data augmentations on sample selection, we conducted an ablation study using the HPH dataset, as shown in 
Table~\ref{tab_data_aug}.
Among the spatial transforms, the highest accuracy was achieved when combining 
random crop, random rotation, and random flip, resulting in an improvement from 
0.645 (no augmentation) to 0.684. When only the color transform 
(color jitter) was applied, the accuracy improved more significantly, 
reaching
0.867. 
This is mainly due to that pathological images are sensitive to color variations, 
while being invariant to spatial transformations.
Finally, combining both spatial and color transforms achieved the 
highest accuracy of 0.904, 
suggesting the importance of the combination of spatial and color transformations for our MVC. }

\subsubsection{Hyper-parameter study}
Our method has three core hyper-parameters: $K$, $M$ and $\gamma$ that represent the
augmentation times, ratio (\%) of selection for $\mathcal{D}_{ir}$ and confidence threshold in HCS, respectively. 
Fig.~\ref{fig:hyper} shows that a larger $K$ results in a higher 
ACC, until it reaches a plateau 
when $K=20$. A larger $M$ obtains more selected samples, but 
with decreased quality.
An elbow of ACC on selected samples is observed when $M$ = 30.
Therefore, we set $M$ = 30 for 
trade-off between quality of pseudo-labels and the selected sample number.
Fig.~\ref{fig:hyper}(c) shows that  $\gamma=0.5$ (i.e., CPS~\cite{cps}) obtained the lowest performance, and the model performed the best 
when $\gamma=0.8$.

\begin{table*}
    \centering
    \caption{Comparison with zero-shot inference by different VLMs for patch-level classification.
    Bold and underlined indicate the best and second-best values, respectively.
    LoRA: Low-Rank Adaptation. LP: Linear Probing. FT: Fine-Tune. FSL: Fully-Supervised Learning. The ACC and Recall are the same on LC25K due to the same number of samples for different classes.}
    \begin{tabular}{c|ccc|ccc|ccc}
    \hline
    \multirow{2}*{Method}& \multicolumn{3}{c|}{~~HPH~~} & \multicolumn{3}{c|}{~~LC25K~~} & \multicolumn{3}{c}{~~NCT-CRC-HE-100K~~}\\ \cline{2-10}
    &  ~~ACC~~ & ~~~~F1~~~ & Recall & ~~ACC~~ & ~~F1~~ & ~~Recall~~ & ~~ACC~~ & ~~F1~~ & ~~Recall~~\\ \hline
    PLIP ~\cite{plip} & 0.645 & 0.636 & 0.746 & 0.709 & 0.704 & 0.709  & 0.501 & 0.510 & 0.506\\
    CLIP ~\cite{clip} & 0.409 & 0.409 & 0.540 & 0.357 & 0.329 & 0.357 & 0.234 & 0.137 & 0.194  \\ 
    BioCLIP ~\cite{bioclip} & 0.607 & 0.588 & 0.650 & 0.169 & 0.094 & 0.169 & 0.171& 0.097 & 0.152\\
    BioMedCLIP ~\cite{biomedclip}  & 0.734 & 0.689 & 0.720 & 0.634 & 0.643 & 0.634 & 0.609 &0.573 & 0.598 \\
    PubMedCLIP~\cite{pubmedclip} & 0.742 & 0.426 & 0.500 & 0.212 & 0.136 & 0.212 & 0.335 & 0.193 & 0.296  \\ 
    Ensemble of~\cite{plip} and \cite{biomedclip} & 0.758 & 0.721 & 0.762 & 0.698 &0.700 & 0.698 & 0.666 & 0.665 & 0.664 \\ \hline
    Ours (UNI-LoRA) & \textbf{0.883} & \textbf{0.856} & \textbf{0.881} & \textbf{0.971} & \textbf{0.971} & \textbf{0.971} & \textbf{0.936}& \textbf{0.936} & \textbf{0.936} \\ 
    Ours (ResNet50-FT) & \underline{0.871} & \underline{0.836} & \underline{0.845} & \underline{0.951} & \underline{0.950} & \underline{0.951} & \underline{0.887} & \underline{0.883} & \underline{0.882} \\ \hline
    FSL (UNI-LoRA) & 0.949 & 0.934 & 0.941 & 0.993 & 0.993 & 0.993 & 0.982 & 0.982 & 0.983 \\ 
    FSL (ResNet50-FT) & 0.917 & 0.895 & 0.906 & 0.973 & 0.972 & 0.973 & 0.974 & 0.974 & 0.975 \\
    FSL (PLIP-LP) & 0.893 & 0.860 & 0.861 & 0.976 & 0.975 & 0.976 & 0.952 & 0.952 & 0.951 \\
    {FSL (BioMedCLIP-FT)} & {0.917} & {0.888} & {0.898} & {0.991} & {0.991} & {0.991} & {0.978} & {0.977} & {0.977} \\
    {FSL (PLIP-FT)} & {0.932} & {0.905} & {0.905} & {0.995} & {0.995} & {0.995} & {0.992} & {0.992} & {0.991} \\ \hline
    \end{tabular}
    \label{tab:zero-shot}
\end{table*}
\subsubsection{Comparison with direct inference with VLMs}
To show the superiority of our method to direct zero-shot inference with VLMs,
we compared our VLM-CPL with five state-of-the-art pre-trained VLMs that were used for zero-shot inference, 
where four of them
are tailored for medical imaging: 1) PLIP~\cite{plip}; 2) CLIP~\cite{clip};
3) BioCLIP~\cite{bioclip} that is trained on the 10M biological image-text pairs;
4) BioMedCLIP~\cite{biomedclip};
5) PubMedCLIP~\cite{pubmedclip} that is a fine-tuned version of CLIP~\cite{clip} based on PubMed articles.

For our method, with the patch-level pseudo-labels, we compared two architectures for training $\phi_A$ and $\phi_B$:
1) UNI-LoRA refers to using the UNI~\cite{UNI} encoder as the backbone and updating 
the weight parameters by LoRA~\cite{hu2021lora}; and 2) ResNet50-FT means fine-tuning ResNet50 
pre-trained on ImageNet.
Besides, to understand the gap between our method and Fully Supervised Learning (FSL), we considered {five}
variants of FSL: 
1) UNI-LoRA; 
2) ResNet50-FT;
3) PLIP-LP indicates the use of the PLIP~\cite{plip} image encoder as the backbone, with weight parameters updated 
through linear probing;
{4) PLIP-FT and  5) BioMedCLIP-FT that mean the image encoder of PLIP~\cite{plip} and BioMedCLIP~\cite{biomedclip}  is used as the backbone, respectively, followed by fully connected layers and fine-tuned by updating all model parameters on the target dataset.} 

Quantitative evaluation of them is shown
in Table~\ref{tab:zero-shot}.
On the HPH dataset, among VLMs that are directly used for inference, CLIP obtained the lowest performance, which is mainly due to the large domain shift
between the pre-training datasets and the downstream dataset. PLIP obtained an ACC and Recall of 0.645  and 0.746, respectively. Without any manual annotations, our  VLM-CPL utilizing pseudo-labels derived from PLIP achieved an 
accuracy of 0.883, an F1-score of 
0.856, and an Recall of 0.881, which outperformed PLIP by 23.8, 22.0, and 23.5 percentage points, respectively.
Meanwhile, our method showed comparable performance to FSL 
(PLIP-LP), with a slightly 
lower accuracy but a higher Recall.
{Compared to BioMedCLIP-FT and PLIP-FT, our method performed lower by 3.4 and 4.9 percentage points, respectively.}
It can also be observed that using UNI outperformed ResNet50 for implementing $\phi_A$ and $\phi_B$ in our method.

\begin{table}[t]
\centering
\caption{Comparison between our method and clustering methods for annotation-free classification
on the HPH dataset.}
\begin{tabular}{c|ccc}
\hline
Methods & ACC & F1 & Recall \\ \hline
Kmean++ (PLIP backbone)~\cite{plip} & 0.745 & 0.710 & 0.755  \\ 
Kmean++ (CLIP backbone)~\cite{clip} & 0.696 & 0.666 & 0.719     \\ 
Kmean++ (ResNet50-in1k)~\cite{resnet} & 0.607 & 0.588 & 0.650  \\
Kmean++ (ViT-B/16-in21k)~\cite{vit2021}  & 0.744 & 0.702 & 0.733 \\
Kmean++ (UNI backbone)~\cite{UNI}  & \underline{0.790} & 
\underline{0.772} & \underline{0.854} \\
Proposed VLM-CPL & \textbf{0.883} & \textbf{0.856} & \textbf{0.881}   \\ \hline
\end{tabular}
\label{tab.kmeans}
\end{table}

\begin{table}[t]
\centering
\caption{\textcolor{black}{Comparison with different VLMs for patch-level classification on the small and imbalanced CRC-P dataset. The classifier is 
ResNet50-FT.}}
\resizebox{0.49\textwidth}{!}{
\begin{tabular}{c|ccc|ccc}
\hline
    \multirow{2}*{\textcolor{black}{Method}}& \multicolumn{3}{c|}{\textcolor{black}{pseudo-label quality}}  & \multicolumn{3}{c}{\textcolor{black}{Testing performance}}  \\ \cline{2-7} 
        & \textcolor{black}{ACC}& \textcolor{black}{F1} & \textcolor{black}{Recall} & \textcolor{black}{ACC} & \textcolor{black}{F1} & \textcolor{black}{Recall} \\ \hline
\textcolor{black}{PLIP ~\cite{plip}} & \textcolor{black}{0.584} & 
\textcolor{black}{0.520} & \textcolor{black}{0.511}  & \textcolor{black}{0.498} & 
\textcolor{black}{0.501} & \textcolor{black}{0.509} \\ 
\textcolor{black}{CLIP ~\cite{clip}} & \textcolor{black}{0.149} & 
\textcolor{black}{0.104} & \textcolor{black}{0.197}  & \textcolor{black}{0.245} & 
\textcolor{black}{0.121} & \textcolor{black}{0.199}  \\ 
\textcolor{black}{BioMedCLIP ~\cite{biomedclip}} & \textcolor{black}{0.496} & 
\textcolor{black}{0.588} & \textcolor{black}{0.615}  & \textcolor{black}{0.646} & \textcolor{black}{0.603} & \textcolor{black}{0.639}\\
\textcolor{black}{Ensemble of~\cite{plip} and \cite{biomedclip}} & 
\textcolor{black}{\underline{0.616}} & 
\textcolor{black}{\underline{0.632}} & \textcolor{black}{\underline{0.683}} & 
\textcolor{black}{\underline{0.704}} & \textcolor{black}{\underline{0.698}} & \textcolor{black}{\underline{0.710}} \\ 
\hline
\textcolor{black}{VLM-CPL} & \textcolor{black}{\textbf{0.893}} & \textcolor{black}
{\textbf{0.850}} & \textcolor{black}{\textbf{0.843}} & \textcolor{black}
{\textbf{0.841}} & \textcolor{black}{\textbf{0.830}} & \textcolor{black}
{\textbf{0.834}}   \\\hline
\end{tabular}}
\label{tab.crc-p}
\end{table}

On the LC25K dataset,
PLIP outperformed the other VLMs for zero-shot inference, with an accuracy of 0.709.
Our method achieved an average accuracy of 0.971, largely outperforming the PLIP by 26.2 percentage points,
and the performance is close to fully supervised fine-tuning of ResNet50.
For the NCT-CRC-HE-100K dataset, BioMedCLIP achieved the highest ACC of 0.609 among existing VLMs, and the 
ensemble of BioMedCLIP and PLIP further improved it to 0.666.
The proposed VLM-CPL obtained a much higher ACC of 0.936, which largely outperformed zero-shot inference methods.

\begin{table*}[t]
    \centering
    \caption{Comparison with state-of-the-art noisy label learning methods for path-level classification. 
    The network is ResNet50-FT.}
    \begin{tabular}{c|ccc|ccc|ccc}
    \hline
    \multirow{2}*{Method}& \multicolumn{3}{c|}{~~HPH~~} & \multicolumn{3}{c|}{~~LC25K~~} 
    & \multicolumn{3}{c}{~~NCT-CRC-HE-100K~~} \\ \cline{2-10}
    &  ~~ACC~~ & ~~F1~~ & Recall & ~~ACC~~ & ~~F1~~ & ~~Recall~~ & ~~ACC~~ & ~~F1~~ & ~~Recall~~ \\ \hline
    Baseline (retrain) & 0.759 & 0.703 & 0.718 & 0.842 & 0.834 & 0.842 & 0.777 & 0.771 & 0.771 \\ \hline
    Co-teaching~\cite{coteaching} & 0.799 & 0.768 & \underline{0.813} & \underline{0.917} & \underline{0.918} & \underline{0.917} & 0.799 & 0.797 & 0.797 \\ 
    Co-teaching+~\cite{co-teaching+} & 0.833 & 0.739 & 0.709 & 0.855 & 0.856 & 0.855 & 0.782 & 0.775 & 0.774 \\ 
    DivideMix~\cite{dividemix} & 0.827 & 0.783 & 0.797 & 0.879 & 0.879 & 0.879 & 0.785 & 0.783 & 0.778 \\
    ELR~\cite{ELR}  & 0.815 & 0.763 & 0.768 & 0.856 & 0.849 & 0.856 & \underline{0.816} & \underline{0.816} & \underline{0.817} \\
    HAMIL~\cite{zhong2023hamil} & \underline{0.850} & \underline{0.792} & 0.776 & 0.863 & 0.864 & 0.863 
    & 0.787 & 0.774 & 0.775 \\ \hline
    Ours & \textbf{0.871} & \textbf{0.836} & \textbf{0.845}  & \textbf{0.951} & \textbf{0.950} & \textbf{0.951} & \textbf{0.887} & \textbf{0.883} & \textbf{0.882} \\ \hline
    \end{tabular}
    \label{tab.NLL}
\end{table*}



\begin{table*}[t]
    \centering
    \caption{{Comparison with our proposed VLM-CPL and its variant VLM-CPL$^\diamond$ where MVC is replaced by CMVC.``Pseudo-label" and``Testing"  denote results for pseudo-label quality on the selected training samples and model performance on the testing set after training, respectively. }}
    \begin{tabular}{c|c|ccc|ccc|ccc}
    \hline
    \multirow{2}*{{Stage}}&\multirow{2}*{{Method}}& \multicolumn{3}{c|}{~~{HPH}~~} & \multicolumn{3}{c|}{~~{LC25K}~~} & \multicolumn{3}{c}{~~{NCT-CRC-HE-100K}~~}\\ \cline{3-11}
    & &  ~~{ACC}~~ & ~~~~{F1}~~~ & {Recall} & ~~{ACC}~~ & ~~{F1}~~ & ~~{Recall}~~ & ~~{ACC}~~ & ~~{F1}~~ & ~~{Recall}~~\\ \hline
    \multirow{2}*{{Pseudo-label}}&{VLM-CPL} & {\textbf{0.926}} & {\textbf{0.912}} & {\textbf{0.937}} & {\textbf{0.974}} & {\textbf{0.973}} & {\textbf{0.974}} & {\textbf{0.941}} & {\textbf{0.938}} & {\textbf{0.935}} \\
    &{VLM-CPL$^\diamond$} & {0.876} & {{0.876}} & {{0.887}} & {{0.972}} & {{0.971}} & {{0.972}} & {{0.925}} & {{0.921}} & {{0.916}} \\ \hline
    \multirow{4}*{{Testing}}&{VLM-CPL (UNI-LoRA)} & {\textbf{0.883}} & {\textbf{0.856}} & {{0.881}} & {{0.971}} & {{0.971}} & {{0.971}} & {{0.936}} & {{0.936}} & {{0.936}} \\ 
    &{
    VLM-CPL$^\diamond$(UNI-LoRA)} & {{0.860}} & {{0.832}} & {\textbf{0.884}} & {\textbf{0.983}} & {\textbf{0.982}} & {\textbf{0.983}} & {\textbf{0.944}} & {\textbf{0.942}} & {\textbf{0.944}}
    \\ \cline{2-11}
    &{VLM-CPL (ResNet50-FT)} & {{0.871}} & {{0.836}} & {{0.845}} & {{0.951}} & {{0.950}} & {{0.951}} & {{0.887}} & {{0.883}} & {{0.882}} \\
    &{
    VLM-CPL$^\diamond$(ResNet50-FT)} & {{0.802}} & {{0.771}} & {{0.836}} & {{0.953}} & {{0.952}} & {{0.953}} & {{0.897}} & {{0.892}} & {{0.893}} \\ \hline
    \end{tabular}
    \label{tab:mvc-v2}
\end{table*}

\begin{table}[t]
\centering
\caption{Ablation study on the TCGA-RCC dataset for WSI-level classification. CLAM-based aggregator was used to
obtain WSI-level predictions on the testing set.}
\resizebox{0.49\textwidth}{!}{
\begin{tabular}{c|ccc|ccc}
\hline
\multirow{2}*{Method}& \multicolumn{3}{c|}{pseudo-label quality}  & \multicolumn{3}{c}{Testing performance}  \\ \cline{2-7} 
        & ACC & F1 & Recall & ACC & F1 &  Recall \\ \hline
Baseline  & 0.662 & 0.658 & 0.661 & 0.672 & 0.674 & 0.671 \\
+ OSP & 0.679 & 0.675 & 0.678 & 0.719 & 0.719 & 0.718\\
+ OSP + MVC  & 0.719 & 0.717 & 0.717 & 0.775 & 0.778 & 0.775 \\
+ OSP + PFC & 0.715 & 0.714 & 0.715 & 0.766 & 0.764 & 0.765 \\ 
+ OSP + MVC + PFC & \multirow{1}*{0.727} & \multirow{1}*{{0.724}} & \multirow{1}*{{0.725}} & 0.794 & 0.792 & 0.793 \\ \hline
+ OSP + MVC + PFC + CPS & \underline{0.731} & \underline{0.730} & \underline{0.730} & \underline{0.803} & \underline{0.793} & \underline{0.801} \\
+ OSP + MVC + PFC + HCS & \textbf{0.743} & \textbf{0.742} & \textbf{0.745} & \textbf{0.822} & \textbf{0.815} & \textbf{0.830} \\ \hline
\end{tabular}}
\label{tab.ablation_wsi}
\end{table}

\subsubsection{Comparison with cluster-based annotation-free 
methods}
For alternative annotation-free methods, we consider 
unsupervised 
clustering by K-means++~\cite{arthur2007kmeans++} 
on the HPH dataset. Note that for evaluation purpose, we 
assigned the class label manually after clustering with two 
clusters. 
Table~\ref{tab.kmeans} shows the results of clustering with four 
different image feature extractors: PLIP~\cite{plip}, 
CLIP~\cite{clip}, 
ResNet50-in1k, ViT-B/16-in21k and UNI~\cite{UNI}.
It can be observed that using the pre-trained UNI for feature 
extraction and clustering is more effective than the other 
feature extractors, and it obtained an accuracy of 79.0\%. 
In contrast, our VLM-CPL obtained a large improvement with an 
accuracy of 88.1\%, which shows the superiority of our approach 
over clustering-based unsupervised classification methods when 
no human annotations are provided at all.
It's worth mentioning that the clustering-based methods cannot 
determine 
the class label for each cluster automatically, especially for 
multi-class classification tasks. 
On the contrary, our approach can automatically predict the 
class label for a testing sample with a much higher accuracy. 

\subsubsection{\textcolor{black}{Robustness under low-resource and imbalanced settings}}
\textcolor{black}{To evaluate the robustness of our method in low-resource and imbalanced scenarios, we construct a new training subset from the NCT-CRC-HE-100K dataset, termed CRC-P. Specifically, we randomly sample 8,000 training patches according to the following imbalanced class distribution: ADI (15\%), DEB (6\%), LYM (7\%), MUC (10\%), MUS (8\%), NORM (25\%), STR (20\%), and TUM (9\%). This reflects realistic conditions where certain tissue types are underrepresented. The testing set remains the same as in the original full-data setting to ensure fair comparison.
Compared to the results on the full NCT-CRC-HE-100K dataset, all methods experienced a performance drop, which can be mainly attributed to the reduced size and more skewed distribution of the training data.
As shown in the Table~\ref{tab.crc-p}, 
PLIP~\cite{plip} and BioMedCLIP~\cite{biomedclip} achieved pseudo-label accuracy of 0.584 and 0.496, respectively, while their ensemble reached 0.616. 
The test accuracy was 0.498 for PLIP, 0.646 for BioMedCLIP, and 0.704 for the ensemble, respectively.
In contrast, 
our VLM-CPL method achieved a 
pseudo-label accuracy of 0.893 and a test accuracy of 0.841, demonstrating its effectiveness even under low-resource and class-imbalanced settings.}

\subsubsection{Comparison with existing noisy label learning methods}
With pseudo-labels obtained by VLM for training set, we then 
compared our method with five noisy label learning methods: 
1) Co-teaching~\cite{coteaching} that 
selects clean labels based on low training loss;
2) Co-teaching+~\cite{co-teaching+} that selects samples with inconsistent predictions;
3) ELR~\cite{ELR} that adopts early-learning regularization to prevent memorization of noisy labels;
4) DivideMix~\cite{dividemix} that uses Gaussian mixture model of the loss distribution to distinguish clean and noisy labels;
5) HAMIL~\cite{zhong2023hamil} that uses two networks to 
supervise a third one. They were compared with the baseline (retrain) that means 
using cross entropy for pseudo-label learning, and all these  
methods used $\mathcal{D}_p$ obtained from the
same VLM for training, 
while our method used MVC + PFC combined with HCS for noisy label 
learning.
Note that the main difference between our method and existing NLL 
methods lies in selection of clean (reliable) samples from 
$\mathcal{D}_p$. 
All these methods employed ResNet50 as the patch-level classifier 
for fair comparison.

\begin{figure}[t]
    \centering
    \includegraphics[width=0.49\textwidth]{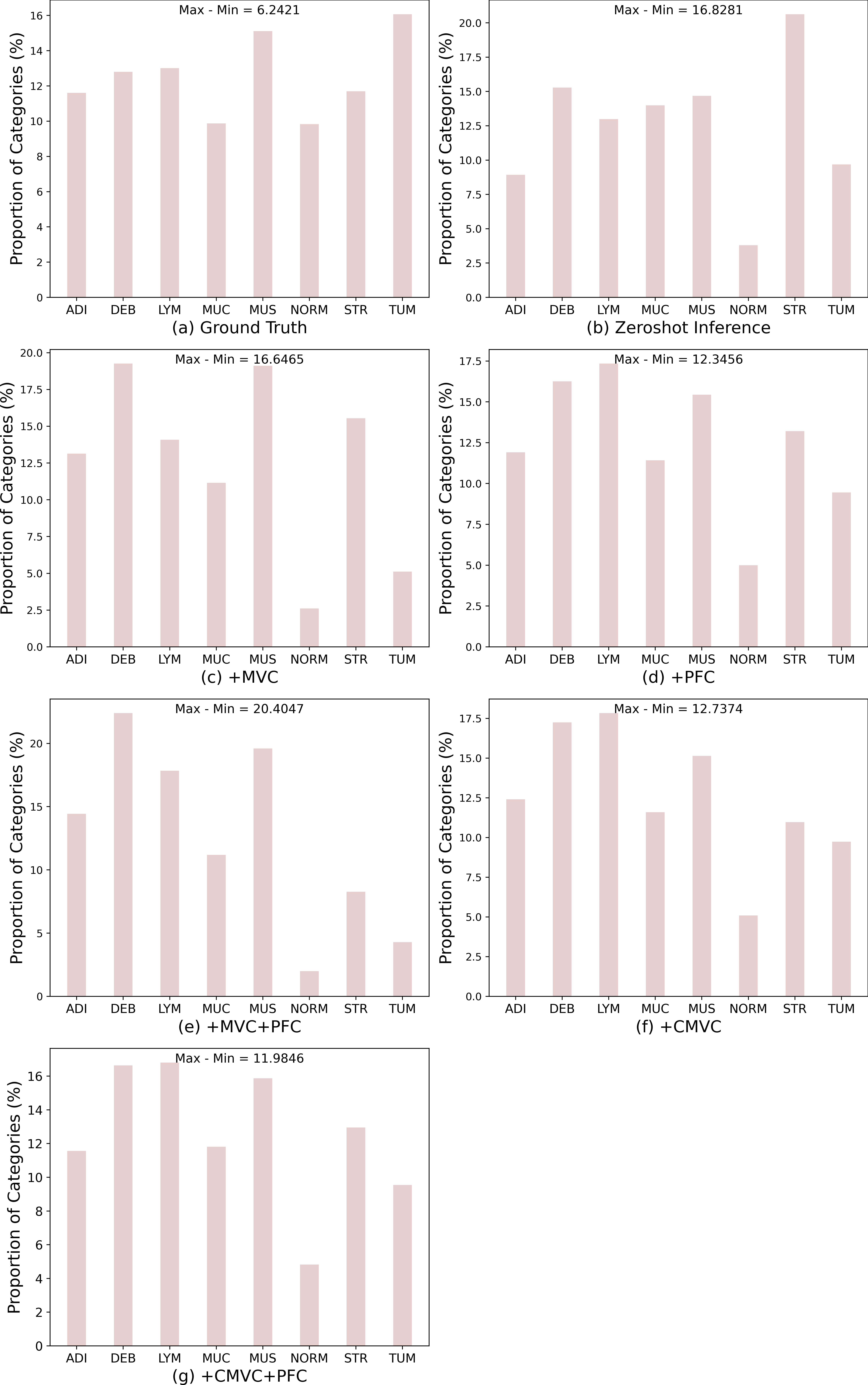}
    \caption{\textcolor{black}{Sample distribution of different filtering strategies on the 
    NCT-CRC-HE-100K dataset.}}
    \label{fig:distribution}
\end{figure}

The results in Table~\ref{tab.NLL} show that the baseline 
obtained the lowest performance on all the datasets. 
For the existing NLL methods, HAMIL~\cite{zhong2023hamil},
Co-teaching~\cite{coteaching} and ELR~\cite{ELR} obtained the 
highest ACC on the three datasets, respectively. 
Our proposed VLM-CPL achieved a better 
performance in all metrics than the existing methods, with an accuracy of 0.871, 0.9510 and 0.887 on the three datasets respectively, which demonstrates that 
VLM-CPL can effectively mitigate the detrimental effects of noisy labels obtained from zero-shot inference of VLMs.

\subsubsection{Effectiveness of CMVC}
{
To analyze the effect of potential class imbalance problem in the selected 
pseudo-labels, we compared the class distribution in selected samples during 
training. The results on the NCT-CRC-100K dataset is shown in 
Fig.~\ref{fig:distribution}. It can be observed that the original training 
dataset is relatively balanced. Using MVC could indeed lead to some class 
imbalance, e.g.,  the ``Norm'' type has a low frequency. Using PFC leads to a 
less imbalance, and MVC + PFC lead to the highest imbalance. In contrast, CMVC 
is better than MVC in terms of class balance, and CMVC + PFC leads to the best 
class balance. 
Furthermore, we compare pseudo-label quality and model performance on testing set between MVC and CMVC in Table~\ref{tab:mvc-v2}. In terms of pseudo-label quality, CMVC is better than MVC on the LC25K dataset while worse on the HPH and 
NCT-CRC-HE-100K datasets. This discrepancy may be due to that CMVC introduces a higher number of unreliable pseudo-labels. 
Despite the slightly lower pseudo-label accuracy of CMVC on the NCT-CRC-HE-100K dataset, the performance of the trained model surpasses using MVC. Table~\ref{tab:mvc-v2} also shows that CMVC outperforms MVC on LC25K and NTC-CRC-HE-100K datasets in terms of testing performance on different backbone networks. Despite the slight discrepancy between MVC and CMVC, both variants of our method outperformed existing methods. Therefore, we still use MVC in the following experiments.}

\begin{figure}[t]
    \centering
    \includegraphics[width=0.48\textwidth]{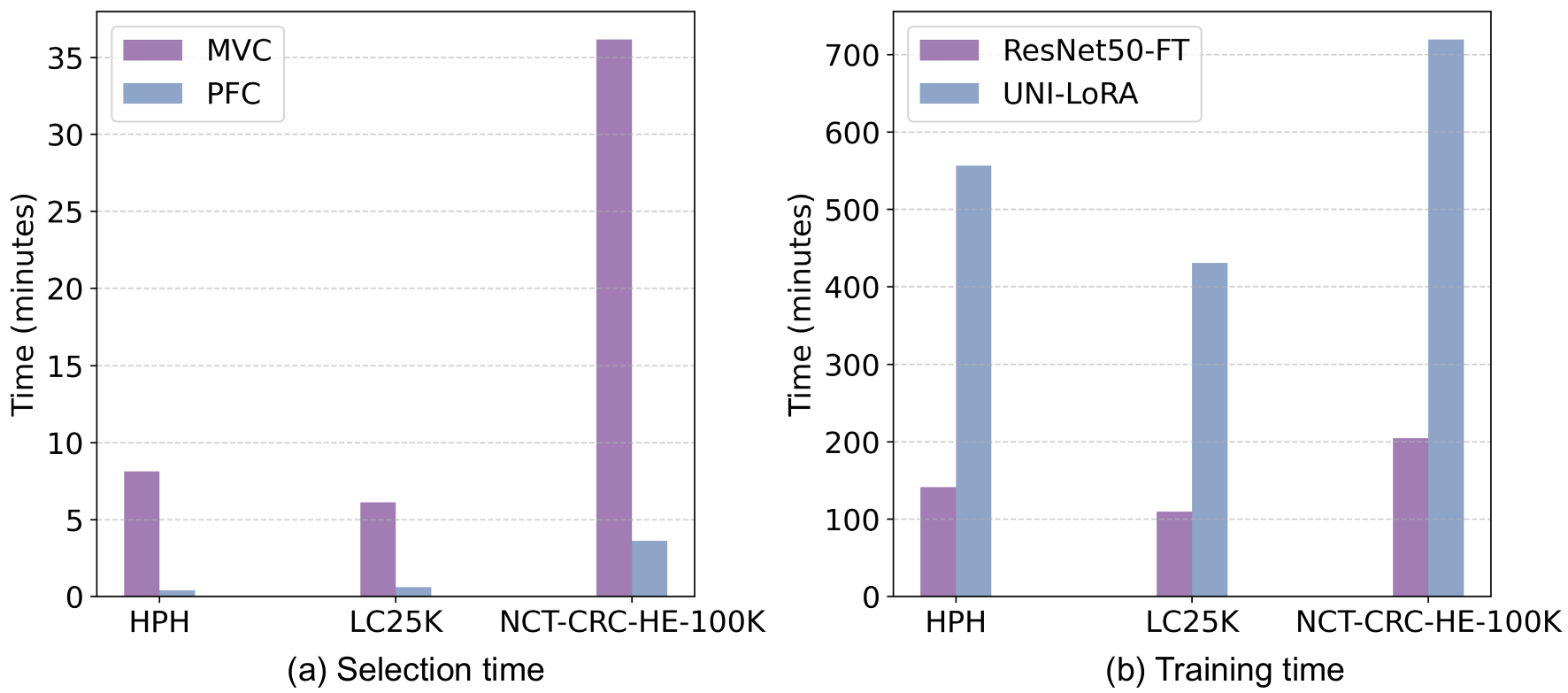}
    \caption{{Total time consumption (minutes) for sample selection and model training on the three patch-level datasets. 
    }}
    \label{fig.time}
\end{figure}
\subsubsection{{Training time and efficiency}}
{To investigate the efficiency of VLM-CPL, we analyzed the time consumed by sample selection and model training, respectively. The results on three datasets are shown in Fig.~\ref{fig.time}. 
It can be observed that the time spent on sample selection is significantly less than the time required for model training, and the time required for PFC is also less than MVC.
The higher time cost for selection and training on the NCT-CRC-HE-100K dataset was mainly due to that we used an
ensemble of PLIP and BioMedCLIP on this dataset, and it is much larger than the HPH and LC25K datasets. For the HPH, LC25K and NCT-CRC-HE-100K datasets, when using ResNet50 as the classification model,
the total time required by the training stage was 141.55, 109.72 and 204.79 minutes, respectively, showing the efficiency of VLM-CPL.
Replacement of ResNet50-FT with UNI-LoRA increased training time to 556.55, 430.73, and 719.72 minutes, respectively, due to higher computational cost and model parameters. However, there is a corresponding improvement in the performance of the model as well, as shown in Table~\ref{tab:zero-shot}.}




\subsection{Results for WSI-level Classification}

\subsubsection{Ablation study}
The TCGA-RCC dataset was used for ablation study in WSI classification tasks. We first evaluate the quality of 
WSI-level pseudo-labels obtained by 
different strategies of using VLMs, and also report the performance on testing samples when trained with these pseudo-labels by using  CLAM~\cite{CLAM} as the aggregator and UNI-LoRA as the patch classifiers. 
The baseline was direct patch-level zero-shot inference using VLMs and mean pooling-based aggregation for a training WSI. 
Table~\ref{tab.ablation_wsi} shows that the pseudo-label accuracy of the baseline was 0.662,
and OSP improved it to 0.679.
Combining OSP with MVC obtained an ACC of 0.719 on pseudo-labels, and OSP +  MVC + PFC further improved it to 0.727. 
Then, employing HCS further improved the pseudo-label ACC to 0.743, and the corresponding ACC on the testing samples was 0.822, which largely outperformed the baseline by 15 percentage points, and was also better than the other methods for obtaining pseudo-labels for WSIs.

\begin{table}[t]
    \centering
    \caption{Comparison between our method with zero-shot inference of VLMs
    on DigestPath and TCGA-RCC testing sets. OSP is not applied for the DigestPath dataset, given that all patches pertain to the target class.}
    \resizebox{0.49\textwidth}{!}{
    \begin{tabular}{c|ccc|ccc}
    \hline
    \multirow{2}*{Method}& \multicolumn{3}{c|}{DigestPath} & \multicolumn{3}{c}{TCGA-RCC} \\ \cline{2-7}
    &  ACC & F1 & Recall & ACC & F1 & Recall \\ \hline
    PLIP ~\cite{plip} & 0.470 & 0.469 & 0.487 & 0.654 & 0.650 & \underline{0.726}  \\
    CLIP ~\cite{clip} & 0.378 & 0.274 & 0.500 & 0.337 & 0.199 & 0.340   \\ 
    BioCLIP ~\cite{bioclip} & 0.568 & 0.405 & 0.469 & 0.383 & 0.275 & 0.381 \\
    BioMedCLIP ~\cite{biomedclip}  & 0.765 & 0.751 & 0.752 & 0.672 & 0.669 & 0.688 \\
    PubMedCLIP~\cite{pubmedclip} & 0.378 & 0.274 & 0.500 & 0.336 & 0.171 & 0.333  \\ 
    Ensemble of~\cite{plip} and \cite{biomedclip} & 0.757 & 0.744 & 0.746 & 0.682 & 0.676 & 0.719 \\ 
    Ensemble of~\cite{plip} and \cite{biomedclip} + OSP & - & - & - & \underline{0.710} & \underline{0.702} & 0.708 \\ \hline
    Ours (Mean pooling) & \underline{0.841} & \underline{0.829} & \underline{0.825} & 0.729 & 0.728 & 0.728  \\ 
    Ours (CLAM) & \textbf{0.909} & \textbf{0.902} & \textbf{0.900} & \textbf{0.822} & \textbf{0.815} & \textbf{0.830}   \\ \hline
    FSL & 0.969 & 0.967 & 0.968 & 0.953 & 0.952 & 0.953 \\ \hline
    \end{tabular}}
    \label{tab:zero-shot-wsi}
\end{table}

\subsubsection{Comparison with direct inference with VLMs}
We also compared our method with zero-shot inference with VLMs on the testing sets, where the latter methods employed mean pooling 
(Eq.~\ref{eq.mean}) 
to aggregate patch-level predictions to a slide-level prediction. 
We also investigated whether OSP benefits zero-shot inference on the TCGA-RCC dataset. 
The results are presented in Table~\ref{tab:zero-shot-wsi}.
On the DigestPath dataset, among the VLMs, 
BioMedCLIP achieved the highest ACC of 0.765, F1-score of 0.751, and 
recall of 0.752.
On the TCGA-RCC dataset, the ensemble of PLIP and BioMedCLIP achieved the highest ACC of 0.682 and F1-score of 0.676, respectively.
By applying OSP, the average ACC was improved to 0.710.
Our method achieved an ACC of 0.909 and 0.822 on the two datasets, respectively. 
Compared with direct inference by BioMedCLIP~\cite{biomedclip}, 
it improved the ACC by 14.4 and 15.0 percentage points 
on the two datasets, respectively.
Replacing CLAM by mean pooing for the aggregator in our method decreased the accuracy by 6.8 and 9.3 percentage points on two datasets, respectively. 

\section{{Discussion}}
{When human annotations are not provided, generating pseudo-labels via VLMs is promising for training a high-performance classification model for pathological images. However, this is largely challenged by the quality of pseudo-labels and inherent bias in the VLM. \textcolor{black}{Fig.~\ref{fig:distribution} shows that the raw prediction from VLM leads to severe class imbalance though the dataset has a relatively balanced distribution of different classes. This is 
intrinsic to the pre-trained VLM that inherently exhibits bias when applied to an unseen downstream dataset, and such bias is challenging to mitigate, especially without additional human inputs.}
\textcolor{black}{Despite this, our method produces a sufficient number of high-quality pseudo-labels for each class and achieves more balanced class distribution, 
as demonstrated by the comparison between 
Fig.~\ref{fig:distribution}(b) and (g), where the percentage gap between the 
most and least frequent classes is reduced from 16.83 to 11.98. 
In future work, incorporating class-aware calibration 
techniques~\cite{li2024multi} or adaptive sampling 
strategies~\cite{cui2019class} may further improve the performance on minority 
classes in 
low-resource and imbalanced settings.}

Our MVC requires a hyper-parameter $M$ to select samples. For a new dataset, the optimal value of $M$ is unknown as it depends on the characteristics of the dataset and the capacity of the VLM. However, it can be chosen empirically based on the trade-off between the sample number and quality of pseudo-labels after filtering. As shown in Fig.~\ref{fig:MVC}, $M=30$ is effective for all three datasets, which ensures that the selected pseudo-labels have a high accuracy and the number of selected samples is not too small. Automatically determining the value of $M$ based on the distribution of uncertainty could enhance the applicability of our approach to diverse datasets.
\textcolor{black}{In future work, we plan to explore adaptive thresholding strategies based on the distribution of entropy values, which may provide a more flexible way to select high-quality pseudo-labels and achieve a better trade-off between the sample number and quality of pseudo-labels after filtering.}

\begin{table}
\centering
\caption{{Quantitative evaluation of selected pseudo-labels with CLIP on the HPH and NCT-CRC-HE-100K datasets.}}
\label{tab.clip}
\begin{tabular}{c|ccc|ccc}
\hline
\multirow{2}*{{Method}}& \multicolumn{3}{c|}{{HPH}}  & \multicolumn{3}{c}{{NCT-CRC-HE-100K}}  \\ \cline{2-7} 
     & {$N_s$}& {ACC} & {F1} & {$N_s$} & {ACC} & {F1} \\ \hline
{Baseline} & {\textbf{17127}} & {0.416} & {0.413} & {71547} & {0.238} & {0.138} \\
{+ MVC} & {5137} & {0.439} & {0.438} & {21464} & {0.309} & {0.157} \\
{+ PFC} & {8888} & {0.591} & {0.582} & {21299} & {0.446} & {0.289} \\
{+ MVC + PFC} & {2749} & {\textbf{0.662}} & {\textbf{0.661}} & {\textbf{6389}} & {\textbf{0.528}} & {\textbf{0.293}} \\ \hline
\end{tabular}
\end{table}

{To further investigate the effectiveness of our VLM-CPL when the 
vision-language model lacks specific knowledge for downstream tasks, we replaced the pathology VLM with CLIP~\cite{clip} that is pretrained on natural images and does 
not contain specific knowledge of the downstream task. For MVC, we used the parameters that were previously mentioned, and Table~\ref{tab.clip} presents the experimental results on the HPH and NCT-CRC-HE-100K datasets.
It can be seen that using CLIP for zero-shot inference alone resulted in the worst
performance, with accuracy rates of 0.416 and 0.238, respectively. Using MVC alone improved the 
accuracy to 0.439 and 0.309, while using PFC alone raised it to 0.591 and 0.446 on the two datasets, respectively. The 
best results were achieved when both MVC and PFC were used together, obtaining an
accuracy of 0.662 and 0.528 on the two datasets, respectively. 
This indicates that despite CLIP not containing specific knowledge of the downstream
task, our proposed method can still improve the quality of pseudo-labels. It is worth noting that the quality of  pseudo-labels obtained by CLIP is much lower those those generated by PLIP~\cite{plip} or BioMedCLIP~\cite{biomedclip}. This further highlights that 
using a VLM with some domain knowledge of pathological images, combined with the two proposed 
filtering methods, can yield more promising results.}

{VLM-CPL may struggle to handle difficult or corner cases, which are likely filtered out 
by MVC and PFC. 
These cases often lie near the decision boundary, where samples tend to be more uncertain or inconsistent, making them harder to classify accurately.
However, the underlying reason for this limitation 
lies in the inherent capabilities of the VLM itself. Pre-trained vision-language models 
often lack the ability to effectively process challenging cases, as they rely solely on 
the information they have learned during pre-training. Consequently, VLM-CPL performs 
well on simpler cases where sufficient knowledge is available but may struggle with 
more complex or ambiguous examples. This issue emphasizes the need for further research 
to explore ways of integrating domain-specific knowledge or human feedback~\cite{qu2023openal,ICAL}, which could 
help improve the model’s performance on these difficult cases.}

{Note that AL~\cite{ICAL} methods also select samples based on uncertainty, and our method has several key differences from them: 1) AL relies on a ``human-in-the-loop'' process, selecting unlabeled samples for manual annotation to iteratively improve model performance, while 
VLM-CPL is fully automated, leveraging pre-trained vision-language models to generate pseudo-labels
without human intervention; 2) AL focuses on annotating uncertain cases, whereas 
VLM-CPL prioritizes selecting high-confidence pseudo-labeled samples to ensure the quality of training data.}

\section{Conclusion}
We presented a novel human annotation-free method VLM-CPL for pathological image
classification that leverages a pre-trained VLM to generate pseudo-labels for the training set.
To address the noisy pseudo-labels caused by domain shift between the 
pre-training and downstream datasets, we propose two consensus filtering methods 
to select clean samples for model training.
First, multi-view consensus utilizes entropy of predictions from multiple augmented versions of an input to reject unreliable pseudo-labels. Second, prompt-feature consensus considers the consensus between
prompt-based pseudo-labels and feature-based pseudo-labels to further select reliable ones.
Based on the reliable subset and the remaining samples without labels, we propose high-confidence cross supervision  
for model training. 
The method was extended with an open-set prompting to filter out irrelevant patches for 
WSI-level classification tasks. 
Without human annotations, our method largely improved the performance from direct zero-shot 
inference of VLMs, which shows potential in clinical practice.
In the future, we plan to extend our approach to pathology image segmentation tasks.

\bibliographystyle{IEEEbib}
\bibliography{mybib}

\begin{thebibliography}{10}
\providecommand{\url}[1]{#1}
\csname url@rmstyle\endcsname
\providecommand{\newblock}{\relax}
\providecommand{\bibinfo}[2]{#2}
\providecommand\BIBentrySTDinterwordspacing{\spaceskip=0pt\relax}
\providecommand\BIBentryALTinterwordstretchfactor{4}
\providecommand\BIBentryALTinterwordspacing{\spaceskip=\fontdimen2\font plus
\BIBentryALTinterwordstretchfactor\fontdimen3\font minus \fontdimen4\font\relax}
\providecommand\BIBforeignlanguage[2]{{%
\expandafter\ifx\csname l@#1\endcsname\relax
\typeout{** WARNING: IEEEtran.bst: No hyphenation pattern has been}%
\typeout{** loaded for the language `#1'. Using the pattern for}%
\typeout{** the default language instead.}%
\else
\language=\csname l@#1\endcsname
\fi
#2}}

\bibitem{skrede2020deep}
O.-J. Skrede, S.~De~Raedt, A.~Kleppe, T.~S. Hveem, K.~Liest{\o}l, J.~Maddison, H.~A. Askautrud, M.~Pradhan, J.~A. Nesheim, F.~Albregtsen, \emph{et~al.}, ``Deep learning for prediction of colorectal cancer outcome: a discovery and validation study,'' \emph{The Lancet}, vol. 395, no. 10221, pp. 350--360, 2020.

\bibitem{campanella2019clinical}
G.~Campanella, M.~G. Hanna, L.~Geneslaw, A.~Miraflor, V.~Werneck Krauss~Silva, K.~J. Busam, E.~Brogi, V.~E. Reuter, D.~S. Klimstra, and T.~J. Fuchs, ``Clinical-grade computational pathology using weakly supervised deep learning on whole slide images,'' \emph{Nature medicine}, vol.~25, no.~8, pp. 1301--1309, 2019.

\bibitem{kather2016multi}
J.~N. Kather, C.-A. Weis, F.~Bianconi, S.~M. Melchers, L.~R. Schad, T.~Gaiser, A.~Marx, and F.~G. Z{\"o}llner, ``Multi-class texture analysis in colorectal cancer histology,'' \emph{Scientific reports}, vol.~6, no.~1, p. 27988, 2016.

\bibitem{abmil}
M.~Ilse, J.~Tomczak, and M.~Welling, ``Attention-based deep multiple instance learning,'' in \emph{International conference on machine learning}.\hskip 1em plus 0.5em minus 0.4em\relax PMLR, 2018, pp. 2127--2136.

\bibitem{shao2021transmil}
Z.~Shao, H.~Bian, Y.~Chen, Y.~Wang, J.~Zhang, X.~Ji, \emph{et~al.}, ``Transmil: Transformer based correlated multiple instance learning for whole slide image classification,'' \emph{NeurIPS}, vol.~34, pp. 2136--2147, 2021.

\bibitem{cps}
X.~Chen, Y.~Yuan, G.~Zeng, and J.~Wang, ``Semi-supervised semantic segmentation with cross pseudo supervision,'' in \emph{CVPR}, 2021, pp. 2613--2622.

\bibitem{luo2022urpc}
X.~Luo, G.~Wang, W.~Liao, J.~Chen, T.~Song, Y.~Chen, S.~Zhang, D.~N. Metaxas, and S.~Zhang, ``Semi-supervised medical image segmentation via uncertainty rectified pyramid consistency,'' \emph{Medical Image Analysis}, vol.~80, p. 102517, 2022.

\bibitem{beluch2018power}
W.~H. Beluch, T.~Genewein, A.~N{\"u}rnberger, and J.~M. K{\"o}hler, ``The power of ensembles for active learning in image classification,'' in \emph{Proceedings of the IEEE conference on computer vision and pattern recognition}, 2018, pp. 9368--9377.

\bibitem{clip}
A.~Radford, J.~W. Kim, C.~Hallacy, A.~Ramesh, G.~Goh, S.~Agarwal, G.~Sastry, A.~Askell, P.~Mishkin, J.~Clark, \emph{et~al.}, ``Learning transferable visual models from natural language supervision,'' in \emph{ICML}, 2021, pp. 8748--8763.

\bibitem{plip}
Z.~Huang, F.~Bianchi, M.~Yuksekgonul, T.~J. Montine, and J.~Zou, ``A visual--language foundation model for pathology image analysis using medical twitter,'' \emph{Nature Medicine}, pp. 1--10, 2023.

\bibitem{biomedclip}
S.~Zhang, Y.~Xu, N.~Usuyama, J.~Bagga, R.~Tinn, S.~Preston, R.~Rao, M.~Wei, N.~Valluri, C.~Wong, \emph{et~al.}, ``Large-scale domain-specific pretraining for biomedical vision-language processing,'' \emph{arXiv preprint arXiv:2303.00915}, 2023.

\bibitem{conch}
M.~Y. Lu, B.~Chen, D.~F. Williamson, R.~J. Chen, I.~Liang, T.~Ding, G.~Jaume, I.~Odintsov, L.~P. Le, G.~Gerber, \emph{et~al.}, ``A visual-language foundation model for computational pathology,'' \emph{Nature Medicine}, vol.~30, no.~3, pp. 863--874, 2024.

\bibitem{enhancingclip}
C.~Menghini, A.~Delworth, and S.~Bach, ``Enhancing clip with clip: Exploring pseudolabeling for limited-label prompt tuning,'' \emph{Advances in Neural Information Processing Systems}, vol.~36, pp. 60\,984--61\,007, 2023.

\bibitem{vpt}
M.~Jia, L.~Tang, B.-C. Chen, C.~Cardie, S.~Belongie, B.~Hariharan, and S.-N. Lim, ``Visual prompt tuning,'' in \emph{ECCV}.\hskip 1em plus 0.5em minus 0.4em\relax Springer, 2022, pp. 709--727.

\bibitem{OEEM}
Y.~Li, Y.~Yu, Y.~Zou, T.~Xiang, and X.~Li, ``Online easy example mining for weakly-supervised gland segmentation from histology images,'' in \emph{MICCAI}.\hskip 1em plus 0.5em minus 0.4em\relax Springer, 2022, pp. 578--587.

\bibitem{lin2022pdbl}
J.~Lin, G.~Han, X.~Pan, Z.~Liu, H.~Chen, D.~Li, X.~Jia, Z.~Shi, Z.~Wang, Y.~Cui, \emph{et~al.}, ``Pdbl: Improving histopathological tissue classification with plug-and-play pyramidal deep-broad learning,'' \emph{IEEE Transactions on Medical Imaging}, vol.~41, no.~9, pp. 2252--2262, 2022.

\bibitem{moyes2023multi}
A.~Moyes, R.~Gault, K.~Zhang, J.~Ming, D.~Crookes, and J.~Wang, ``Multi-channel auto-encoders for learning domain invariant representations enabling superior classification of histopathology images,'' \emph{Medical Image Analysis}, vol.~83, p. 102640, 2023.

\bibitem{xue2021selective}
Y.~Xue, J.~Ye, Q.~Zhou, L.~R. Long, S.~Antani, Z.~Xue, C.~Cornwell, R.~Zaino, K.~C. Cheng, and X.~Huang, ``Selective synthetic augmentation with histogan for improved histopathology image classification,'' \emph{Medical image analysis}, vol.~67, p. 101816, 2021.

\bibitem{CLAM}
M.~Y. Lu, D.~F. Williamson, T.~Y. Chen, R.~J. Chen, M.~Barbieri, and F.~Mahmood, ``Data-efficient and weakly supervised computational pathology on whole-slide images,'' \emph{Nature biomedical engineering}, vol.~5, no.~6, pp. 555--570, 2021.

\bibitem{mambamil}
S.~Yang, Y.~Wang, and H.~Chen, ``Mambamil: Enhancing long sequence modeling with sequence reordering in computational pathology,'' \emph{arXiv preprint arXiv:2403.06800}, 2024.

\bibitem{align}
C.~Jia, Y.~Yang, Y.~Xia, Y.-T. Chen, Z.~Parekh, H.~Pham, Q.~Le, Y.-H. Sung, Z.~Li, and T.~Duerig, ``Scaling up visual and vision-language representation learning with noisy text supervision,'' in \emph{International conference on machine learning}.\hskip 1em plus 0.5em minus 0.4em\relax PMLR, 2021, pp. 4904--4916.

\bibitem{mizero}
M.~Y. Lu, B.~Chen, A.~Zhang, D.~F. Williamson, R.~J. Chen, T.~Ding, L.~P. Le, Y.-S. Chuang, and F.~Mahmood, ``Visual language pretrained multiple instance zero-shot transfer for histopathology images,'' in \emph{Proceedings of the IEEE/CVF conference on computer vision and pattern recognition}, 2023, pp. 19\,764--19\,775.

\bibitem{quilt-1M}
W.~Ikezogwo, S.~Seyfioglu, F.~Ghezloo, D.~Geva, F.~Sheikh~Mohammed, P.~K. Anand, R.~Krishna, and L.~Shapiro, ``Quilt-1m: One million image-text pairs for histopathology,'' \emph{NeurIPS}, vol.~36, 2024.

\bibitem{han2018co-teach}
B.~Han, Q.~Yao, X.~Yu, G.~Niu, M.~Xu, W.~Hu, I.~Tsang, and M.~Sugiyama, ``Co-teaching: Robust training of deep neural networks with extremely noisy labels,'' in \emph{NeurIPS}, 2018, pp. 8527--8537.

\bibitem{yu2019co-teaching+}
X.~Yu, B.~Han, J.~Yao, G.~Niu, I.~Tsang, and M.~Sugiyama, ``How does disagreement help generalization against label corruption?'' in \emph{ICML}.\hskip 1em plus 0.5em minus 0.4em\relax PMLR, 2019, pp. 7164--7173.

\bibitem{dividemix}
J.~Li, R.~Socher, and S.~C. Hoi, ``Dividemix: Learning with noisy labels as semi-supervised learning,'' in \emph{ICLR}, 2019, pp. 1--14.

\bibitem{zhong2023hamil}
L.~Zhong, G.~Wang, X.~Liao, and S.~Zhang, ``Hamil: High-resolution activation maps and interleaved learning for weakly supervised segmentation of histopathological images,'' \emph{IEEE Transactions on Medical Imaging}, 2023.

\bibitem{GCE}
Z.~Zhang and M.~Sabuncu, ``Generalized cross entropy loss for training deep neural networks with noisy labels,'' in \emph{NeurIPS}, 2018, pp. 8778--8788.

\bibitem{ELR}
S.~Liu, J.~Niles-Weed, N.~Razavian, and C.~Fernandez-Granda, ``Early-learning regularization prevents memorization of noisy labels,'' \emph{NeurPIS}, vol.~33, pp. 20\,331--20\,342, 2020.

\bibitem{uncertainty-view}
M.~Abdar, F.~Pourpanah, S.~Hussain, D.~Rezazadegan, L.~Liu, M.~Ghavamzadeh, P.~Fieguth, X.~Cao, A.~Khosravi, U.~R. Acharya, \emph{et~al.}, ``A review of uncertainty quantification in deep learning: Techniques, applications and challenges,'' \emph{Information fusion}, vol.~76, pp. 243--297, 2021.

\bibitem{wang2019aleatoric}
G.~Wang, W.~Li, M.~Aertsen, J.~Deprest, S.~Ourselin, and T.~Vercauteren, ``Aleatoric uncertainty estimation with test-time augmentation for medical image segmentation with convolutional neural networks,'' \emph{Neurocomputing}, vol. 338, pp. 34--45, 2019.

\bibitem{gaillochet2022taal}
M.~Gaillochet, C.~Desrosiers, and H.~Lombaert, ``Taal: Test-time augmentation for active learning in medical image segmentation,'' in \emph{MICCAI Workshop on Data Augmentation, Labelling, and Imperfections}.\hskip 1em plus 0.5em minus 0.4em\relax Springer, 2022, pp. 43--53.

\bibitem{wang_tta}
G.~Wang, W.~Li, M.~Aertsen, J.~Deprest, S.~Ourselin, and T.~Vercauteren, ``Aleatoric uncertainty estimation with test-time augmentation for medical image segmentation with convolutional neural networks,'' \emph{Neurocomputing}, vol. 338, pp. 34--45, 2019.

\bibitem{ayhan2018test}
M.~S. Ayhan and P.~Berens, ``Test-time data augmentation for estimation of heteroscedastic aleatoric uncertainty in deep neural networks,'' in \emph{Medical Imaging with Deep Learning}, 2018.

\bibitem{wu2022mutual}
Y.~Wu, Z.~Ge, D.~Zhang, M.~Xu, L.~Zhang, Y.~Xia, and J.~Cai, ``Mutual consistency learning for semi-supervised medical image segmentation,'' \emph{Medical Image Analysis}, vol.~81, p. 102530, 2022.

\bibitem{kmeans++}
D.~Arthur and S.~Vassilvitskii, ``K-means++ the advantages of careful seeding,'' in \emph{ACM-SIAM symposium on Discrete algorithms}, 2007, pp. 1027--1035.

\bibitem{zhu2011group}
H.~Zhu, M.~Zhou, and R.~Alkins, ``Group role assignment via a kuhn--munkres algorithm-based solution,'' \emph{IEEE Transactions on Systems, Man, and Cybernetics-Part A: Systems and Humans}, vol.~42, no.~3, pp. 739--750, 2011.

\bibitem{qu2023openal}
L.~Qu, Y.~Ma, Z.~Yang, M.~Wang, and Z.~Song, ``Openal: An efficient deep active learning framework for open-set pathology image classification,'' in \emph{MICCAI}.\hskip 1em plus 0.5em minus 0.4em\relax Springer, 2023, pp. 3--13.

\bibitem{salvi2021hybrid}
M.~Salvi, M.~Bosco, L.~Molinaro, A.~Gambella, M.~Papotti, U.~R. Acharya, and F.~Molinari, ``A hybrid deep learning approach for gland segmentation in prostate histopathological images,'' \emph{Artificial Intelligence in Medicine}, vol. 115, p. 102076, 2021.

\bibitem{LC25K}
A.~A. Borkowski, M.~M. Bui, L.~B. Thomas, C.~P. Wilson, L.~A. DeLand, and S.~M. Mastorides, ``Lung and colon cancer histopathological image dataset (\textsc{LC}25000),'' \emph{arXiv preprint arXiv:1912.12142}, 2019.

\bibitem{digestpath}
Q.~Da, X.~Huang, Z.~Li, Y.~Zuo, C.~Zhang, J.~Liu, W.~Chen, J.~Li, D.~Xu, Z.~Hu, \emph{et~al.}, ``Digestpath: A benchmark dataset with challenge review for the pathological detection and segmentation of digestive-system,'' \emph{Medical Image Analysis}, vol.~80, p. 102485, 2022.

\bibitem{Sensecare2024}
G.~Wang, Q.~Duan, T.~Shen, and S.~Zhang, ``Sensecare: a research platform for medical image informatics and interactive 3d visualization,'' \emph{Frontiers in Radiology}, vol.~4, pp. 1--18, 2024.

\bibitem{vit2021}
A.~Dosovitskiy, L.~Beyer, A.~Kolesnikov, D.~Weissenborn, X.~Zhai, T.~Unterthiner, M.~Dehghani, M.~Minderer, G.~Heigold, S.~Gelly, J.~Uszkoreit, and N.~Houlsby, ``An image is worth 16x16 words: Transformers for image recognition at scale,'' in \emph{ICLR}, 2021, pp. 1--21.

\bibitem{UNI}
R.~J. Chen, T.~Ding, M.~Y. Lu, D.~F. Williamson, G.~Jaume, A.~H. Song, B.~Chen, A.~Zhang, D.~Shao, M.~Shaban, \emph{et~al.}, ``Towards a general-purpose foundation model for computational pathology,'' \emph{Nature Medicine}, vol.~30, no.~3, pp. 850--862, 2024.

\bibitem{hu2021lora}
E.~J. Hu, P.~Wallis, Z.~Allen-Zhu, Y.~Li, S.~Wang, L.~Wang, W.~Chen, \emph{et~al.}, ``Lora: Low-rank adaptation of large language models,'' in \emph{ICLR}, 2022.

\bibitem{bioclip}
S.~Stevens, J.~Wu, M.~J. Thompson, E.~G. Campolongo, C.~H. Song, D.~E. Carlyn, L.~Dong, W.~M. Dahdul, C.~Stewart, T.~Berger-Wolf, \emph{et~al.}, ``Bioclip: A vision foundation model for the tree of life,'' in \emph{CVPR}, 2024, pp. 19\,412--19\,424.

\bibitem{pubmedclip}
S.~Eslami, G.~de~Melo, and C.~Meinel, ``Does clip benefit visual question answering in the medical domain as much as it does in the general domain?'' \emph{arXiv preprint arXiv:2112.13906}, 2021.

\bibitem{resnet}
K.~He, X.~Zhang, S.~Ren, and J.~Sun, ``Deep residual learning for image recognition,'' in \emph{CVPR}, 2016, pp. 770--778.

\bibitem{coteaching}
B.~Han, Q.~Yao, X.~Yu, G.~Niu, M.~Xu, W.~Hu, I.~Tsang, and M.~Sugiyama, ``Co-teaching: Robust training of deep neural networks with extremely noisy labels,'' in \emph{NeurIPS}, 2018, pp. 8527--8537.

\bibitem{co-teaching+}
X.~Yu, B.~Han, J.~Yao, G.~Niu, I.~Tsang, and M.~Sugiyama, ``How does disagreement help generalization against label corruption?'' in \emph{ICML}, 2019, pp. 7164--7173.

\bibitem{arthur2007kmeans++}
D.~Arthur and S.~Vassilvitskii, ``K-means++ the advantages of careful seeding,'' in \emph{ACM-SIAM}, 2007, pp. 1027--1035.

\bibitem{li2024multi}
S.~Li, L.~Song, X.~Wu, Z.~Hu, Y.-m. Cheung, and X.~Yao, ``Multi-class imbalance classification based on data distribution and adaptive weights,'' \emph{IEEE Transactions on Knowledge and Data Engineering}, pp. 5265--5279, 2024.

\bibitem{cui2019class}
Y.~Cui, M.~Jia, T.-Y. Lin, Y.~Song, and S.~Belongie, ``Class-balanced loss based on effective number of samples,'' in \emph{CVPR}, 2019, pp. 9268--9277.

\bibitem{ICAL}
W.~Hu, L.~Cheng, G.~Huang, X.~Yuan, G.~Zhong, C.-M. Pun, J.~Zhou, and M.~Cai, ``Learning from incorrectness: Active learning with negative pre-training and curriculum querying for histological tissue classification,'' \emph{IEEE Transactions on Medical Imaging}, vol.~43, no.~2, pp. 625--637, 2024.

\end{thebibliography}

\end{document}